\documentclass[lettersize,journal]{IEEEtran}
\usepackage{amsmath,amsfonts}
\usepackage{algorithmic}
\usepackage{algorithm}
\usepackage{booktabs}
\usepackage{array}
\usepackage[caption=false,font=normalsize,labelfont=sf,textfont=sf]{subfig}
\usepackage{textcomp}
\usepackage{stfloats}
\usepackage{graphicx}
\usepackage{subcaption}
\usepackage{caption}
\usepackage{multirow}
\usepackage{multicol}
\usepackage{url}
\usepackage{verbatim}
\usepackage{cite}
\usepackage{xspace}
\usepackage{tikz}
\usepackage{pgfplots} %引用包

\usepackage{scalerel}
\usetikzlibrary{svg.path}
\definecolor{orcidlogocol}{HTML}{A6CE39}
\tikzset{
    orcidlogo/.pic={
        \fill[orcidlogocol] svg{M256,128c0,70.7-57.3,128-128,128C57.3,256,0,198.7,0,128C0,57.3,57.3,0,128,0C198.7,0,256,57.3,256,128z};
        \fill[white] svg{M86.3,186.2H70.9V79.1h15.4v48.4V186.2z}
        svg{M108.9,79.1h41.6c39.6,0,57,28.3,57,53.6c0,27.5-21.5,53.6-56.8,53.6h-41.8V79.1z M124.3,172.4h24.5c34.9,0,42.9-26.5,42.9-39.7c0-21.5-13.7-39.7-43.7-39.7h-23.7V172.4z}
        svg{M88.7,56.8c0,5.5-4.5,10.1-10.1,10.1c-5.6,0-10.1-4.6-10.1-10.1c0-5.6,4.5-10.1,10.1-10.1C84.2,46.7,88.7,51.3,88.7,56.8z};
    }
}
\newcommand\orcidicon[1]{\href{https://orcid.org/#1}{\mbox{\scalerel*{
                \begin{tikzpicture}[yscale=-1,transform shape]
                \pic{orcidlogo};
                \end{tikzpicture}
            }{|}}}}
\usepackage[implicit=false]{hyperref}
\hypersetup{hidelinks,
	colorlinks=true,
	allcolors=black,
	pdfstartview=Fit,
	breaklinks=true}

\newcommand{\name}{PIAT\xspace}

\begin{document}

\title{Parameter Interpolation Adversarial Training for Robust Image Classification}

\author{Xin Liu${\textsuperscript{\orcidicon{0009-0006-7345-2405}}}$, Yichen Yang${\textsuperscript{\orcidicon{0000-0002-6725-8319}}}$, 
Kun He${\textsuperscript{\orcidicon{0000-0001-7627-4604}}}$,~\IEEEmembership{Senior Member,~IEEE}, 
John E. Hopcroft${\textsuperscript{\orcidicon{0000-0001-8681-607}}}$,~\IEEEmembership{Life Fellow,~IEEE}
% \author{IEEE Publication Technology,~\IEEEmembership{Staff,~IEEE}

\thanks{
This work is supported by National Natural Science Foundation of China (U22B2017, 62076105) and International Cooperation Foundation of Hubei Province, China (2024EHA032).}
\thanks{Xin Liu, Yichen Yang and Kun He are with School of Computer Science and Technology, Huazhong University of Scinece and Technology.  
The emails of these authors are liuxin\_jhl@hust.edu.cn, yangyc@hust.edu.cn, brooklet60@hust.edu.cn}% <-this % stops a space
\thanks{ John E. Hopcroft is with Computer Science Department, Cornell University Ithaca, USA.
E-mail: jeh@cs.cornell.edu.}
\thanks{Xin Liu and Yichen Yang contributed equally to this work. The corresponding author is Kun He.}
% \thanks{Manuscript received July 10, 2024; revised ?? ??, 2024.}
%\thanks{Digital Object Identifier xxx/xxxxx}
}

\markboth{IEEE TRANSACTIONS ON INFORMATION FORENSICS AND SECURITY, VOL.xx, NO.xx, JULY 2024}%
{Shell \MakeLowercase{\textit{et al.}}: A Sample Article Using IEEEtran.cls for IEEE Journals}


\maketitle

\begin{abstract}
Though deep neural networks exhibit superior performance on various tasks, they are still plagued by adversarial examples. 
Adversarial training has been demonstrated to be the most effective method to defend against adversarial attacks. 
However, existing adversarial training methods show that the model robustness has apparent oscillations and overfitting issues in the training process, degrading the defense efficacy. 
To address these issues, we propose a novel framework called Parameter Interpolation Adversarial Training (PIAT). 
PIAT tunes the model parameters between each epoch by interpolating the parameters of the previous and current epochs.
It makes the decision boundary of model change more moderate and alleviates the overfitting issue, helping the model converge better and achieving higher model robustness. 
In addition, we suggest using the Normalized Mean Square Error (NMSE) to further improve the robustness by aligning the relative magnitude of logits between clean and adversarial examples rather than the absolute magnitude. 
Extensive experiments conducted on several benchmark datasets demonstrate that our framework could prominently improve the robustness of both Convolutional Neural Networks (CNNs) and Vision Transformers (ViTs).

%TL;DR "Too Long; Didn't Read": a short sentence describing your paper: 
% We propose a novel Parameter Interpolation Adversarial Training (PIAT) framework together with a new regularization term of normalized mean square error, which boosts the model robustness of both Convolutional Neural Networks and Vision Transformers.

%Keywords: Adversarial examples, adversarial training, model robustness
\end{abstract}


\section{Introduction}
Deep Neural Networks (DNNs) have been widely used in various tasks, including computer vision~\cite{resnet,DBLP:conf/nips/KrizhevskySH12,DBLP:conf/cvpr/LongSD15}, natural language processing~\cite{bert,Attention}, and speech recognition~\cite{DBLP:conf/interspeech/SakSRB15,DBLP:conf/interspeech/ChenRLG17}. 
However, they are known to be vulnerable to adversarial examples by injecting malicious perturbations to clean examples that can cause the model to misclassify inputs with high confidence~\cite{FGSM,mahmood2021robustness}. 
Since DNNs have been applied in many safety systems, it is crucial to make them reliable and robust.

\begin{figure}[t]
    \centering
    \includegraphics[width=\columnwidth]{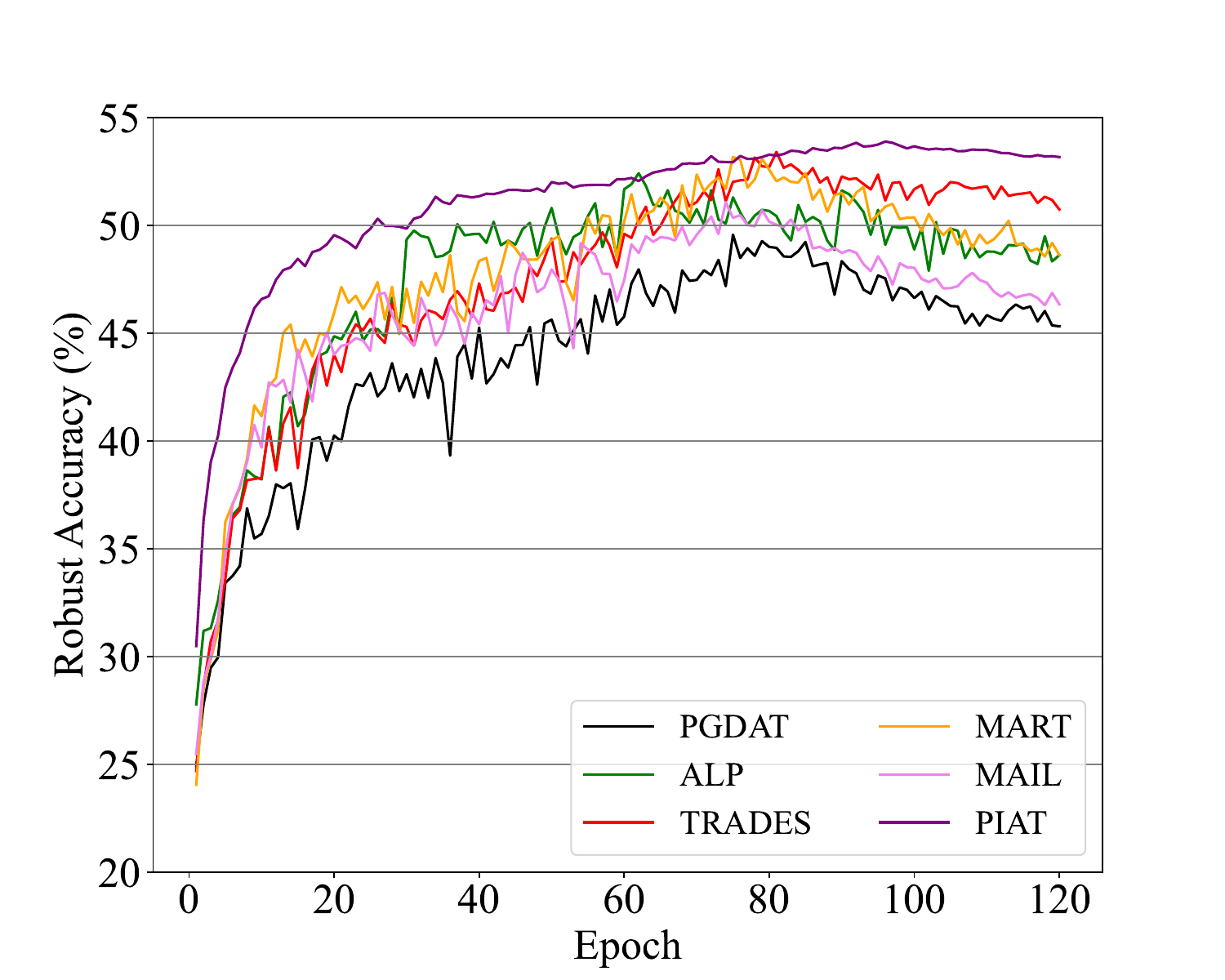}
    \caption{ 
    The robust accuracy of ResNet18 trained on CIFAR10 dataset by existing advanced adversarial training methods has apparent oscillations and overfitting issues in the training process. 
    On the contrary, our PIAT framework achieves excellent robust accuracy with better convergence, further improving the model performance.
    }
    \label{fig:adv_classical}
\end{figure}
As the most effective defense approach, adversarial training dynamically generates adversarial examples and incorporates them during training.
Recently, numerous adversarial training methods have been proposed to boost the model's performance, such as adding regularization term~\cite{ALP,TRADES,8417973,DBLP:journals/pami/YangXHBHCH23,DBLP:journals/tifs/MaXJZS23,DBLP:journals/tifs/LiuLSFC22}, assigning different weights to the data points~\cite{MAIL,DBLP:journals/tifs/LinLCCW23} and adapting to generate suitable adversarial examples~\cite{DBLP:conf/cvpr/LeeLY20,FAT,10478545,DBLP:journals/tifs/YuanZWW23}. 
However, the model robustness remains unsatisfactory due to hard convergence and generalization of adversarial training. 

Previous works~\cite{TRADES,EMAT} have shown that adversarial training yields a more complex decision boundary than standard training. 
Moreover, we observe that the robust accuracy of the model has apparent oscillations in the early training stage, as illustrated in Fig.~\ref{fig:adv_classical} 
Worse still, in the later training stage, the experiments show that the overfitting issue occurs. 
The training accuracy continues to increase, but the robust accuracy of the testing data begins to decline.
Consequently, a natural intuition is that the model robustness can be improved if the adversarial training converges stably without overfitting.

Based on this motivation, we introduce a novel framework called Parameter Interpolation Adversarial Training (PIAT) to solve the apparent oscillations and overfitting issues of model robustness in the training process.
Specifically, PIAT tunes the model parameters between each epoch by interpolating the model parameters of the previous and current epochs. 
To balance the effect of previous accumulated and current parameters, PIAT gradually increases the weight of the previous model parameters when tuning the current model parameters since the model parameters become more valuable during the course of training. 
In other words, PIAT focuses more on current model parameters in the early training stage to make the model converge more stably. 
In the later training stage, PIAT focuses more on previously accumulated parameters, preventing the decision boundary from becoming too complex and alleviating the overfitting issue.

Moreover, there have been many works~\cite{ALP,TRADES,MART} proposed to encourage similarity between the output of clean and adversarial examples.
Particularly, we observe that ALP~\cite{ALP} uses the mean square error loss to align the absolute magnitude of logits between clean and adversarial examples.
However, the data distribution of clean and adversarial examples is quite different, and simply forcing the output to be close is too demanding.
Therefore, we propose a new metric called Normalized Mean Square Error (NMSE) to align the clean and adversarial examples better.
It pays more attention to aligning the relative magnitude rather than the absolute magnitude of logits.

Our main contributions are summarized as follows:
\begin{itemize}
\item To mitigate the oscillations and the overfitting issues in the training process, we propose the PIAT framework that interpolates the model parameters of the previous and current epochs. 
PIAT tunes the model parameters to converge stably, alleviates overfitting issues, and achieves higher robustness.
\item We suggest using NMSE as a new regularization term to better align the clean and adversarial examples. 
NMSE pays more attention to the relative magnitude of the output of clean and adversarial examples rather than the absolute magnitude.
\item 
Extensive experiments demonstrate that our method is an effective and general framework, achieving excellent robustness on both Convolutional Neural Networks (CNNs) and Vision Transformers (ViTs). 
\end{itemize}

\section{Related Work}
\subsection{Adversarial Example}
Let \(\boldsymbol{x}_i\) and \(y_i\) denote a clean example and the corresponding ground-truth label in dataset \(\mathcal{D}={(\boldsymbol{x}_i, y_i)}^n_{i=1}\), where \(\boldsymbol{x}_i \in \mathcal{X}\) and \(y_i \in \mathcal{Y}=\{1,...,c\}\).
The goal of adversaries is to find an adversarial example \(\boldsymbol{x}_i^{adv} \in \mathcal{B}_{\epsilon}[\boldsymbol{x}_i]=\{\boldsymbol{x}_i^{adv}|\|\boldsymbol{x}_i^{adv}-\boldsymbol{x}_i\|_{\infty} \leq \epsilon\}\), which causes errors in the model prediction.

Existing adversarial attacks can be categorized into white-box~\cite{FGSM} and black-box attacks~\cite{MIM}.
In general, the adversarial attacks can achieve good performance in the white-box setting, where the attacker can access the complete information of the target model, including the architecture and model weights.

Numerous methods have been proposed to improve the performance of adversarial examples in the white-box setting.
The Fast Gradient Sign Method (FGSM)~\cite{FGSM} is the first white-box attack that crafts adversarial examples by utilizing the sign of the gradient direction, formulated by:
\begin{equation}
    \boldsymbol{x}^{adv} =\boldsymbol{x}+\epsilon \cdot \operatorname{sign}(\nabla_{\boldsymbol{x}} \mathcal{L}(f_{\boldsymbol{\theta}}(\boldsymbol{x}),y)),
\end{equation}
where $\boldsymbol{x}$ denotes the adversarial example, \(\operatorname{sign}(\cdot)\) is the signal function, and $\epsilon$ is the perturbation magnitude.
Iterative Fast Gradient Sign Method (I-FGSM)~\cite{IFGSM} extends to an iteration version, which generates adversarial examples with multiple iterations and a smaller step size.
Momentum Iterative Fast Gradient Sign Method (MI-FGSM)~\cite{MIM} introduces momentum to enhance the transferability of adversarial examples as follows:
% \textcolor{blue}{
\begin{gather}
\begin{split}
&\boldsymbol{g}_t = \mu \cdot \boldsymbol{g}_{t-1} 
+ \frac{\nabla_{\boldsymbol{x}}(\mathcal{L}(f_{\boldsymbol{\theta}}(\boldsymbol{x}^{adv}_{t-1}),y))}
{\|\nabla_{\boldsymbol{x}}(\mathcal{L}(f_{\boldsymbol{\theta}}(\boldsymbol{x}^{adv}_{t-1}),y)\|_1}, \\
&\boldsymbol{x}^{adv}_t = \boldsymbol{x}^{adv}_{t-1} 
+ \alpha \cdot \operatorname{sign}(\boldsymbol{g}_t).
\end{split}
\end{gather}
% }

where $\boldsymbol{x}_i^t$ denotes the adversarial example at the $t^{th}$ step, \(\boldsymbol{g}_t\) is the gradient at the $t^{th}$ step, \(\|\cdot\|_1\) is the 1-norm  and $\alpha$ is the step size.
Projected Gradient Descent (PGD)~\cite{PGDAT} generates stronger adversarial perturbation by means of multi-step iterative projection, formulated by:
\begin{equation}
    \boldsymbol{x}_i^{t+1}=\prod_{\mathcal{B}_{\epsilon}[\boldsymbol{x}_i]}(\boldsymbol{x}_i^{t}+\alpha \cdot \operatorname{sign}( \nabla_{\boldsymbol{x}_i^t}\mathcal{L}(f_{\boldsymbol{\theta}}(\boldsymbol{x}_i^{t}),y_i)),  
\end{equation}
where $\boldsymbol{x}_i^t$ denotes the adversarial example at the $t^{th}$ step and $\prod(\cdot)$ is the projection operator, 
% \textcolor{blue}{
which constrains the magnitude of the perturbation range.
% }
The C\&W attack~\cite{CW}, which generates adversarial examples by the optimization-based method, is widely used to evaluate the model robustness.
AutoAttack (AA)~\cite{AA} is a parameter-free ensemble attack, which has been widely adopted as one of the criteria for evaluating the model robustness.

\subsection{Adversarial Training}
Adversarial training (AT) has been demonstrated to be the most effective method to defend against adversarial examples.
It dynamically generates adversarial examples and incorporates them during training to improve the robustness of DNNs.
To achieve this goal, PGD-AT~\cite{PGDAT} formulates the adversarial training optimization problem as the following min-max problem:
\begin{equation}
\min_{\boldsymbol{\theta}}\sum_{i}\max_{\boldsymbol{x}^{adv}_i\in \mathcal{B}_{\epsilon}[\boldsymbol{x}_i]}
\mathcal{L}(f_{\boldsymbol{\theta}}(\boldsymbol{x}^{adv}_i),y_i),
\end{equation}
where $f_{\boldsymbol{\theta}}(\cdot): \mathbb{R}^d \rightarrow \mathbb{R}^c$ is the DNN classifier with parameter ${\boldsymbol{\theta}}$. $\mathcal{L}(\cdot, \cdot)$ represents the cross entropy loss.

PGD-AT introduces a new optimization paradigm, and along this multi-step paradigm, numerous works have been explored to further alleviate the adversarial vulnerability of DNNs.
Adding regularization terms, such as ALP~\cite{ALP} and TRADES~\cite{TRADES}, provides a systematic way to better align the logits between clean and corresponding adversarial examples.
MART~\cite{MART} and MMA~\cite{MMA} explicitly differentiate the misclassified and correctly classified examples during  training. 
% \textcolor{blue}{
RAT~\cite{RAT} further adds random noise to deterministic weights and using Taylor expansion, aiming to improve robustness against adversarial examples.
% }
To better utilize the model capacity, weighted adversarial training methods, such as GAIRAT~\cite{GAIRAT} and MAIL~\cite{MAIL}, introduce a weighting strategy where the larger weight is assigned to more vulnerable pointers closer to the decision boundary.
To achieve a better trade-off between robustness and accuracy, some methods, including LBGAT~\cite{LGBAT} and HAT~\cite{HAT}, use the clean example output of the normally trained model to modify the adversarial example output of the adversarial trained model.
MLCAT~\cite{MLCAT}, UIAT~\cite{UIAT}, and STAT~\cite{STAT} focus on maximizing the likelihood of both adversarial examples and neighbouring data points.
ARD and PRM~\cite{ARDPRM} is the first work, which proposes using randomly masking gradients from some attention blocks or masking perturbations to improve the adversarial robustness of ViTs.
% \textcolor{blue}{
CFA~\cite{CFA} customizes training configurations for different classes to enhance both robustness and fairness in adversarial training, addressing disparities in robustness among classes.
% }

The works most related to ours are KDSWA~\cite{KDSWA} and ALP~\cite{ALP}.
KDSWA~\cite{KDSWA} introduces SWA~\cite{SWA}, 
% \textcolor{blue}{
which uses random weight average
% } 
to smooth model weights and mitigates the overfitting issue.
% \textcolor{blue}{
Instead of training one model and random ensemble on another like SWA, our PIAT framework interpolates the previous and current model parameters of the same model to achieve a more moderate change in the decision boundary at each epoch and continues to train the model using the interpolated parameters.
% }
Besides, ALP~\cite{ALP} calculates the regularization term using the absolute magnitude of logits with the MSE loss, while our NMSE focuses on aligning the relative magnitude.

\section{Motivation}
In this section, we first construct a synthetic dataset and explore its decision boundary to investigate the solution for the apparent oscillation issue in adversarial training. 
Then, we provide some theoretical analysis in solving the overfitting issue in adversarial training.  
Finally, we rethink the alignment mode of ALP and present a novel regularization to align the logits between clean and adversarial examples.

\begin{figure}[t]
    \centering
    \subfloat[Data distribution in 2D]{
    \includegraphics[width=0.225\textwidth]{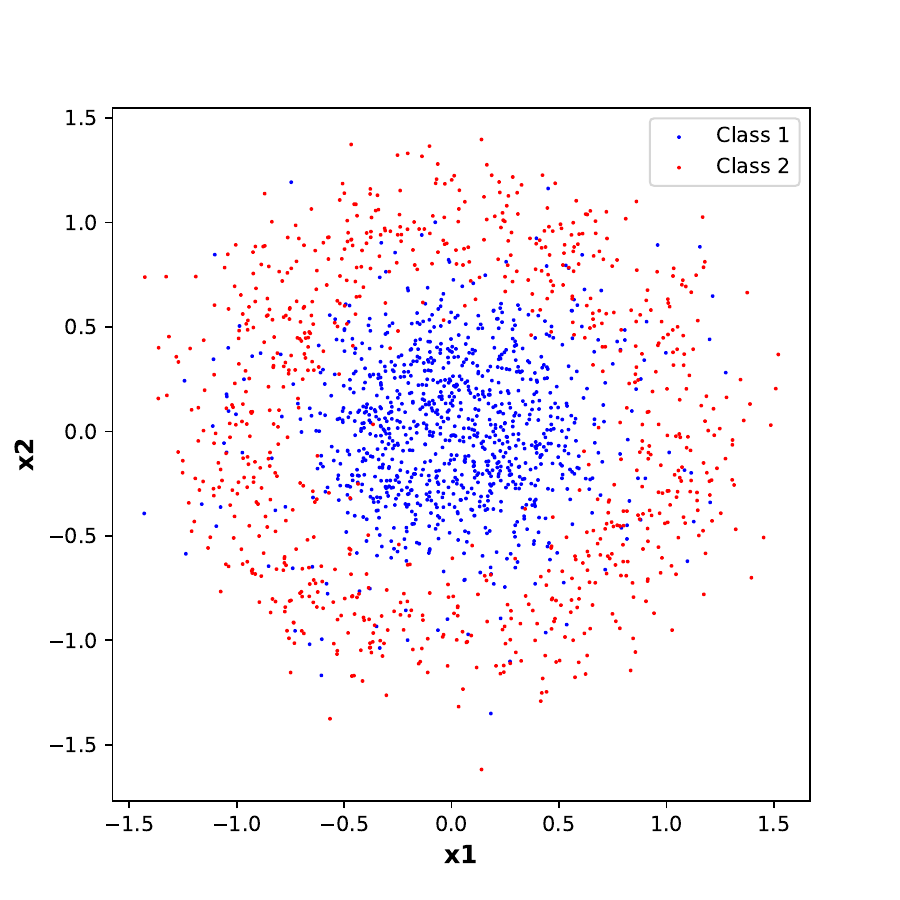}}
    \subfloat[Data distribution in 3D]{
    \includegraphics[width=0.225\textwidth]{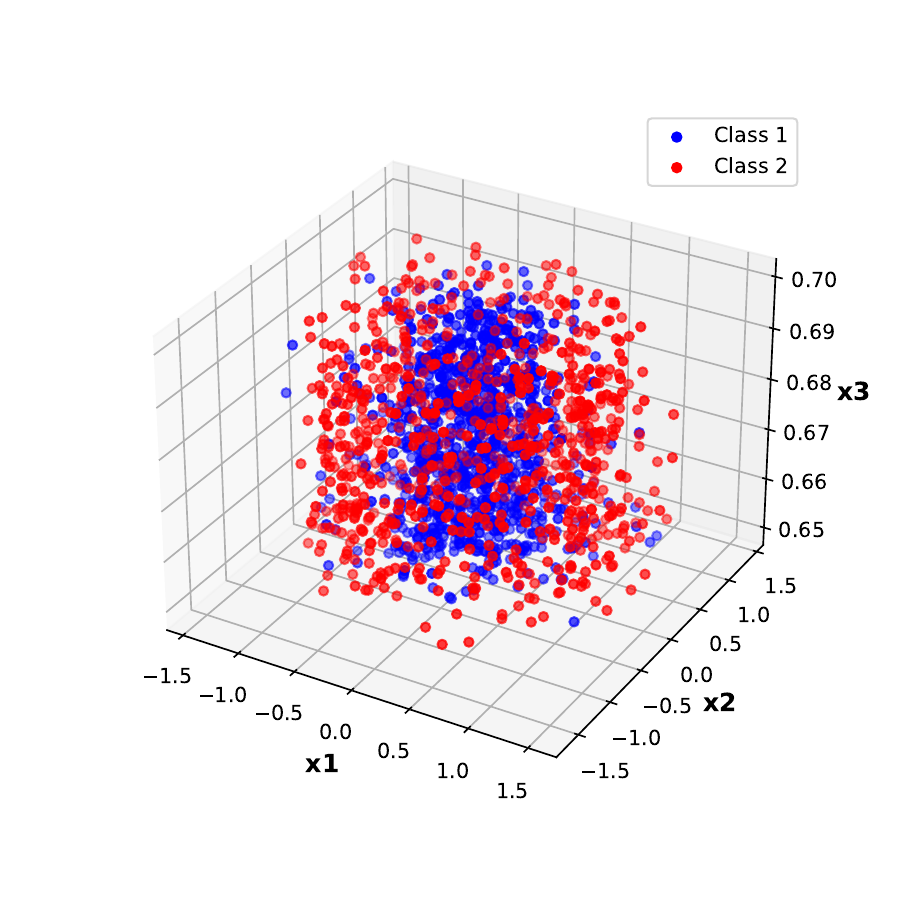}}
    \caption{The data distributions of the toy example, which is two concentric circles with different radii. The class 1 data are primarily located within the inner, while the class 2 data are mainly distributed on the outside.
    }
    \label{fig:Data Distribution}
\end{figure}
\subsection{A Toy Example}
\label{Toy model}
To delve into the techniques of adversarial training, we construct a simple 3D binary classification dataset comprising two distinct data distributions, specifically two concentric circles with different radii, and observe the accuracy and robustness of the model during the training process. 
Figure~\ref{fig:Data Distribution} illustrates the toy dataset in two dimensions (2D) and three dimensions (3D), respectively.
The data used in the toy example comes from two different data distributions, shown by red points and blue points respectively.
Specfically, we generate the three features \(x_1,x_2\) and \(x_3\) using the following equations:
\(x_1 = \rho_i cos(z) +\epsilon_1, x_2 = \rho_i sin(z) + \epsilon_2, x_3 \sim U(\alpha_i, \beta_i)\), where \(z \sim U(0, 2\pi)\) and \(\epsilon_1, \epsilon_2 \sim N (0, \sigma_2)\). Here,  \(i=1\) for class 1 and \(i=2\) for class 2.
We set the parameters as follows: \(\sigma = 0.2, \rho_1 = 0.35, \rho_2 = 1, \alpha_1 = \alpha_2 = 0.80\) and \(\beta_1 = \beta_2 = 0.85\).
We use a single hidden layer of MLP as the training model. For the model training, we use SGD with a momentum of 0.9 and a learning rate to 0.5.
The learning rate value is chosen to reflect the convergence difficulty observed when training on other datasets such as CIFAR10 and CIFAR100.
We train the model through adversarial training for 50 epochs and generate adversarial examples using a PGD attack.
The attack parameters are set as follows: a step size of \(\alpha = 0.05\), a maximum perturbation boundary of \(\epsilon = 0.1\), and iterations \(K=5\) for the adversarial training.
% Detailed information regarding this synthetic dataset is described in Appendix~\ref{Toy Model Settings}. 

\begin{figure*}[t]
    \centering
    \subfloat[The model trained by PGD-AT and PIAT]{
    \includegraphics[width=0.45\textwidth]{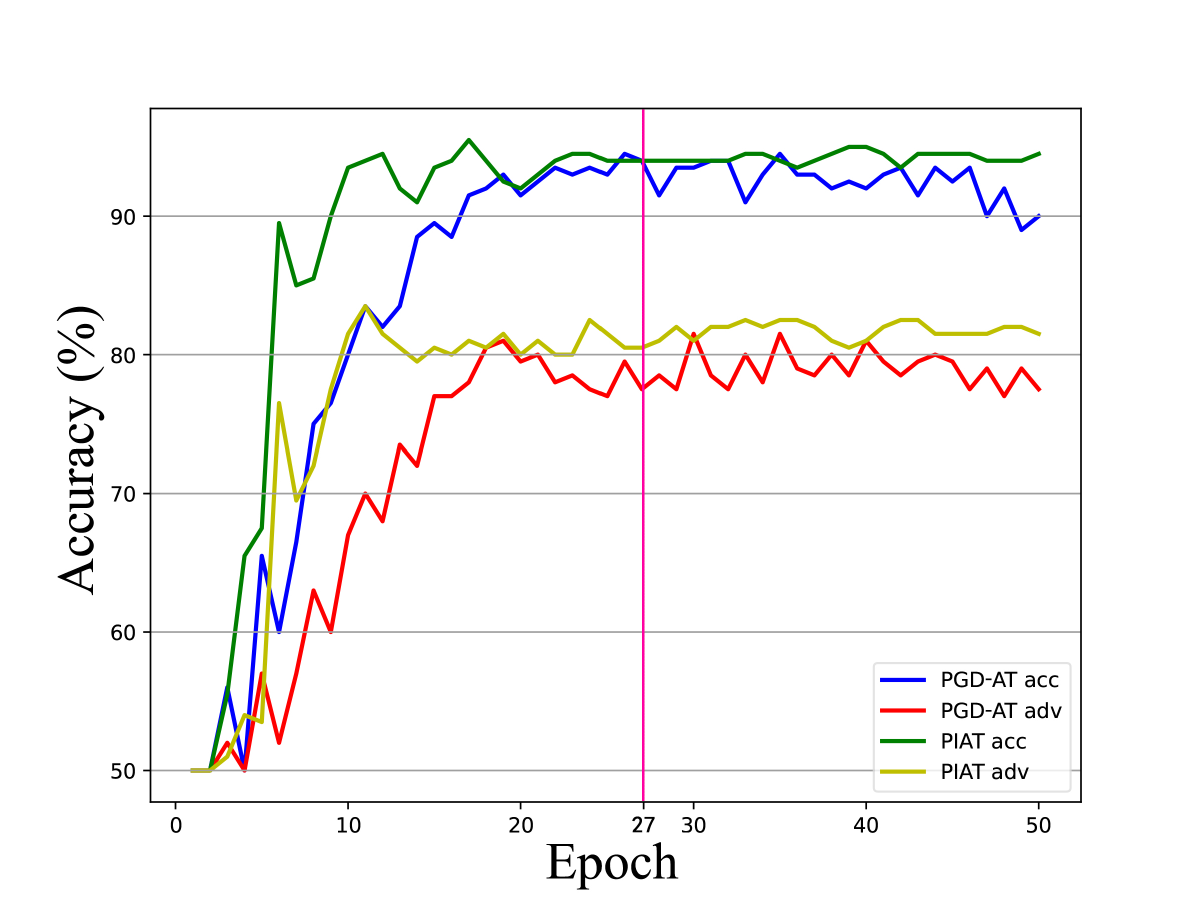}}
    \subfloat[The model trained by ALP and NMSE]{
    \includegraphics[width=0.45\textwidth]{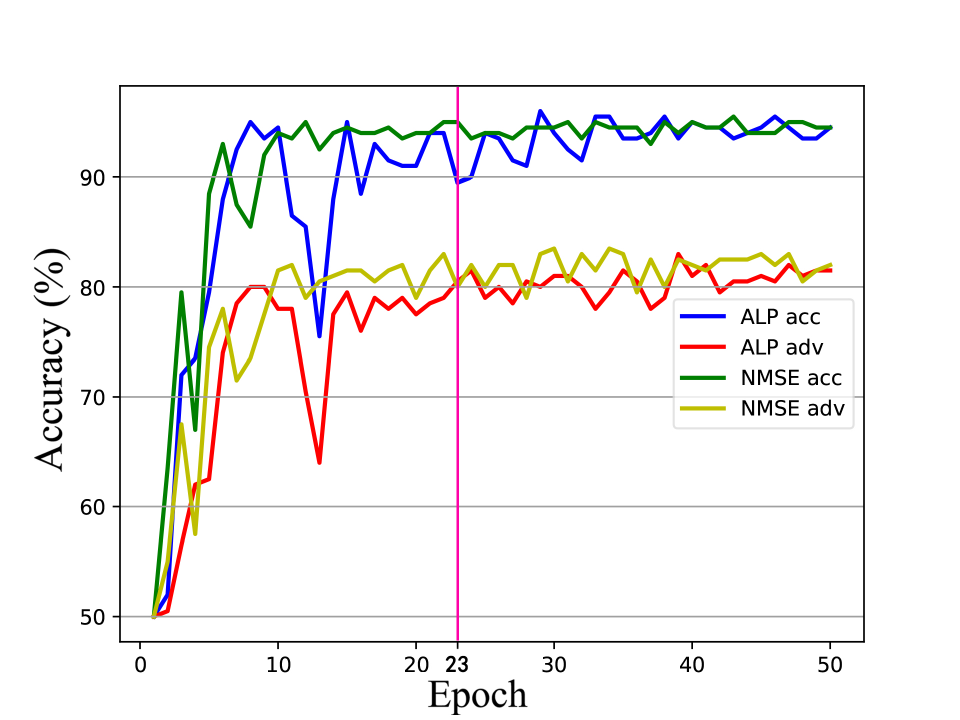}}
    \caption{Illustrations of defense performance under PGD adversarial attack.The first figure illustrates the accuracy and robustness of the toy model trained using PGD-AT and PIAT on the 3D dataset, while the second figure demonstrates the accuracy and robustness of the toy model with ALP and NMSE regularization on the same dataset.
    }
    \label{fig:Accuracy and Robustness}
\end{figure*}
\begin{figure*}[t]
    \centering
    \subfloat[The decision boundary before the 27\(^{th}\) epoch]{
    \includegraphics[width=0.45\textwidth]{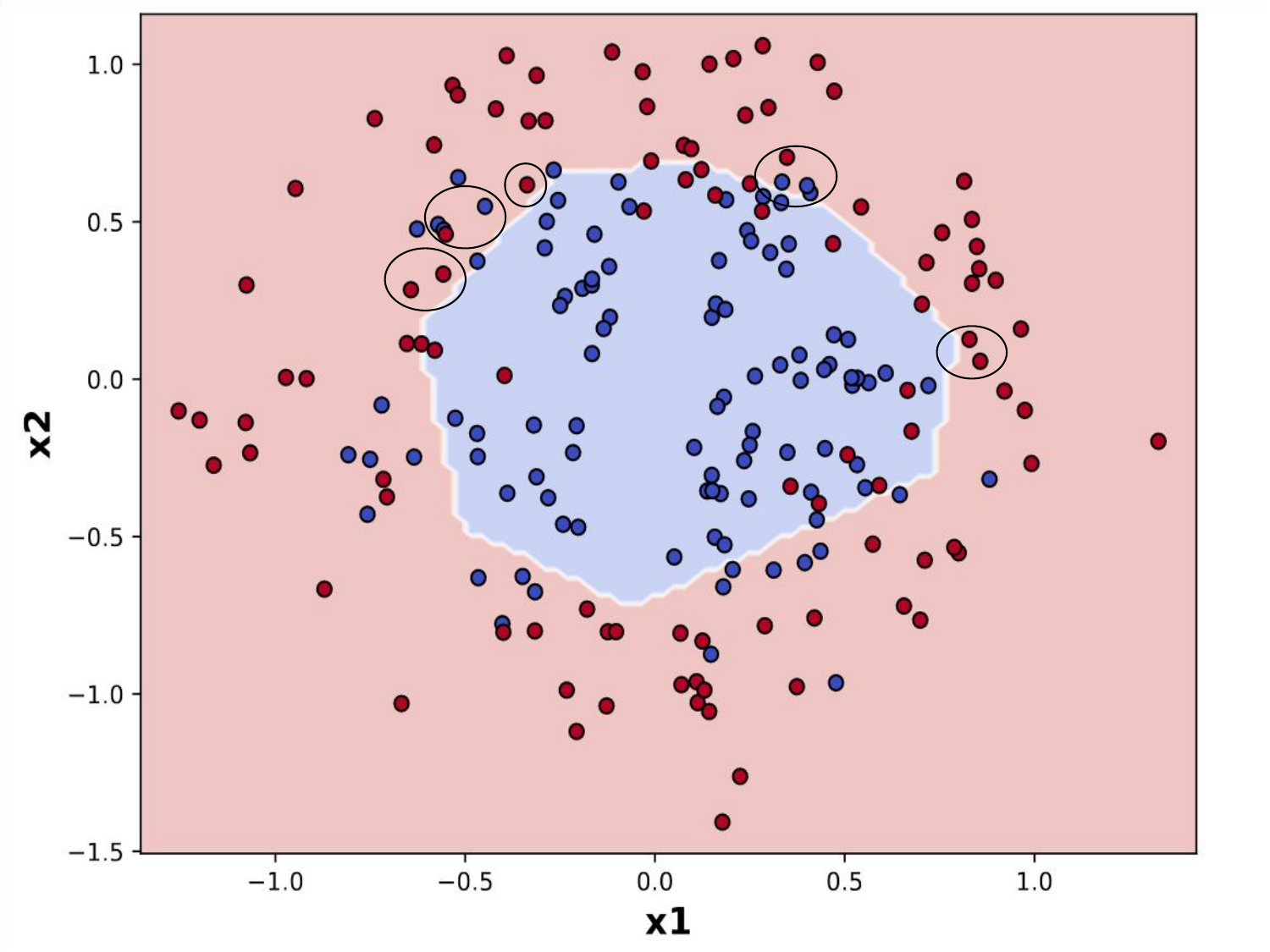}}
    \subfloat[The decision boundary after the 27\(^{th}\) epoch]{
    \includegraphics[width=0.45\textwidth]{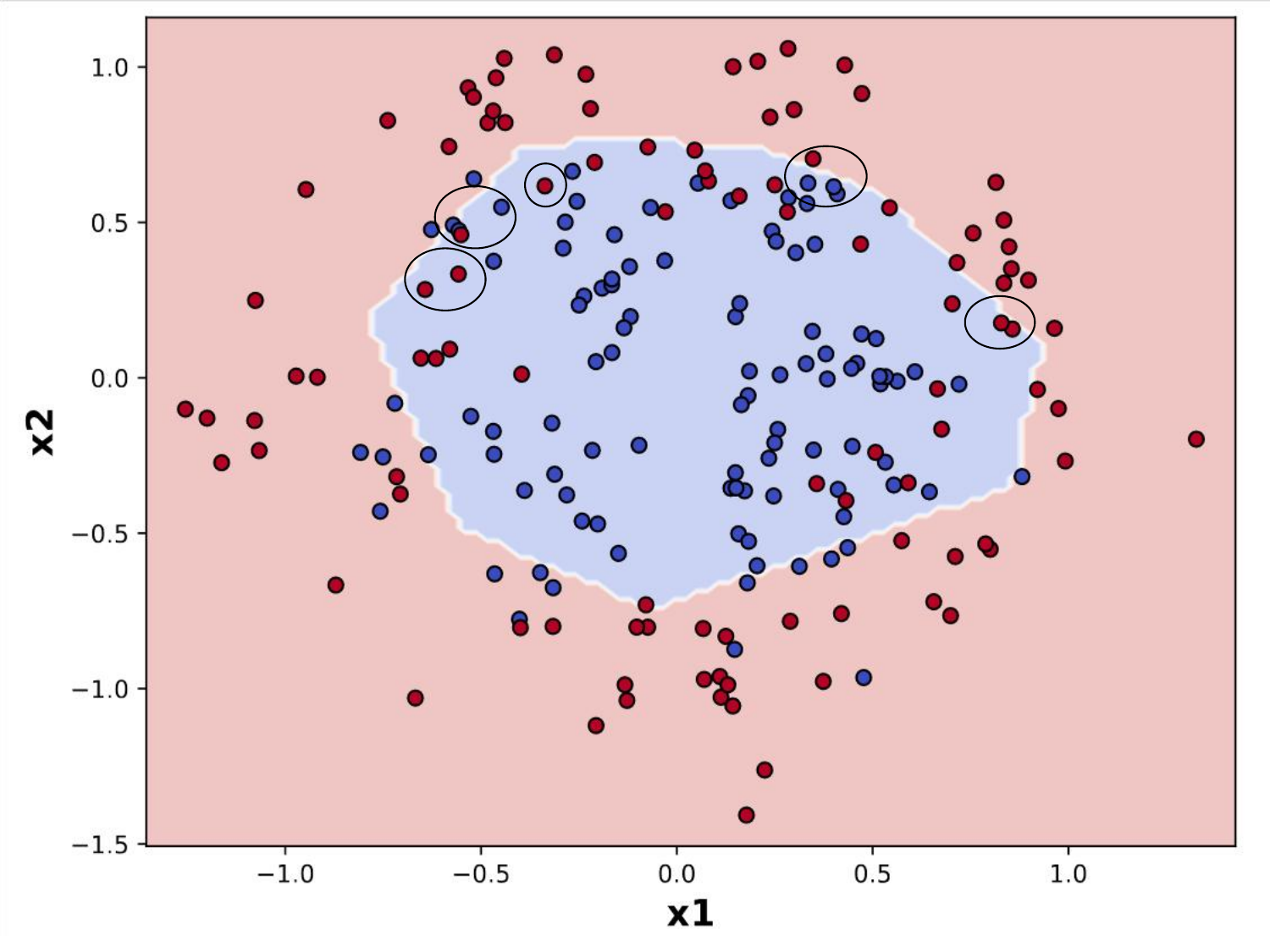}}
    \caption{
    Illustrations of the 2D decision boundary of the model trained using PGD-AT at the 27\(^{th}\) epoch.
    % \textcolor{blue}{
    The corresponding data points are marked by circles.
    % }
    While the blue data points near the top left of the decision boundary are correctly classified, the red data points situated around the top left and right are misclassified.
    }
    \label{fig:Decision Boundary}
\end{figure*}
% \textcolor{blue}{
As illustrated in Fig.~\ref{fig:Accuracy and Robustness}, the robustness of the model trained by PGD-AT exhibits apparent oscillation during training.
To further explore the reason for the oscillation issue, we observe the decision boundary of all the epochs where the robustness exhibits a sudden decrease. 
As shown in Fig.~\ref{fig:Decision Boundary}, taking the 27\(^{th}\) epoch as an example, we observe that the decision boundary of model changes rapidly from the beginning to the end, leading to a significant fluctuation in the model robustness.
% }
However, we could not directly reduce the learning rate of the optimizer because this will slow down the convergence and cause the overfitting issue in the later stage.

The aforementioned phenomenon raises an intriguing question: Can the model achieve improved adversarial robustness by converging more stably when the changes in the decision boundary are relatively smooth?
To implement this idea of mitigating the dramatic change on decision boundary, we tune the model parameters at the end of each epoch by interpolating the model parameters of the previous and current epochs, leading to better initial model parameters for the next epoch. 

\begin{figure*}[t!]
    \centering
    \subfloat[PGD-AT from 26\(^{th}\) to 27\(^{th}\) epoch]{
    \includegraphics[width=0.45\textwidth]{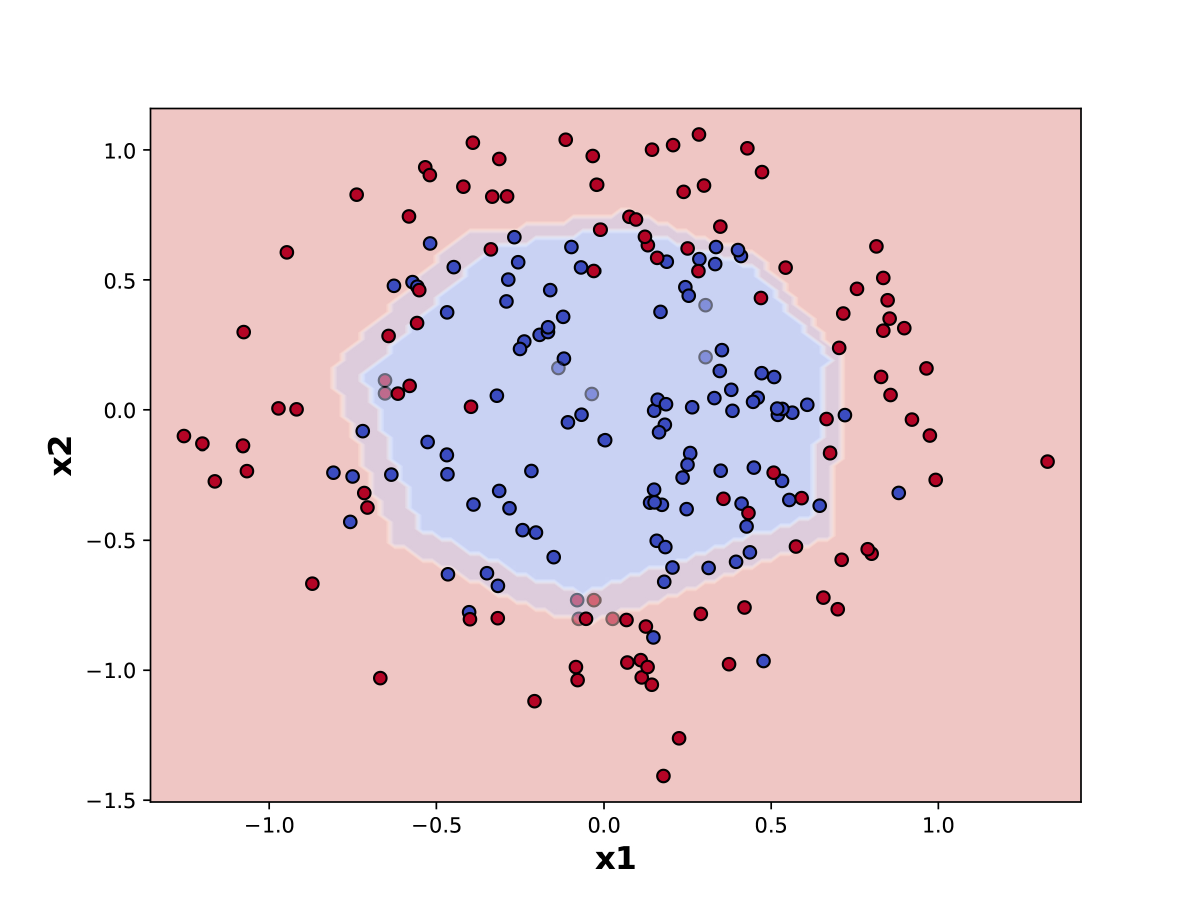}}
    \subfloat[PIAT from 24\(^{th}\) to 25\(^{th}\) epoch]{
    \includegraphics[width=0.45\textwidth]{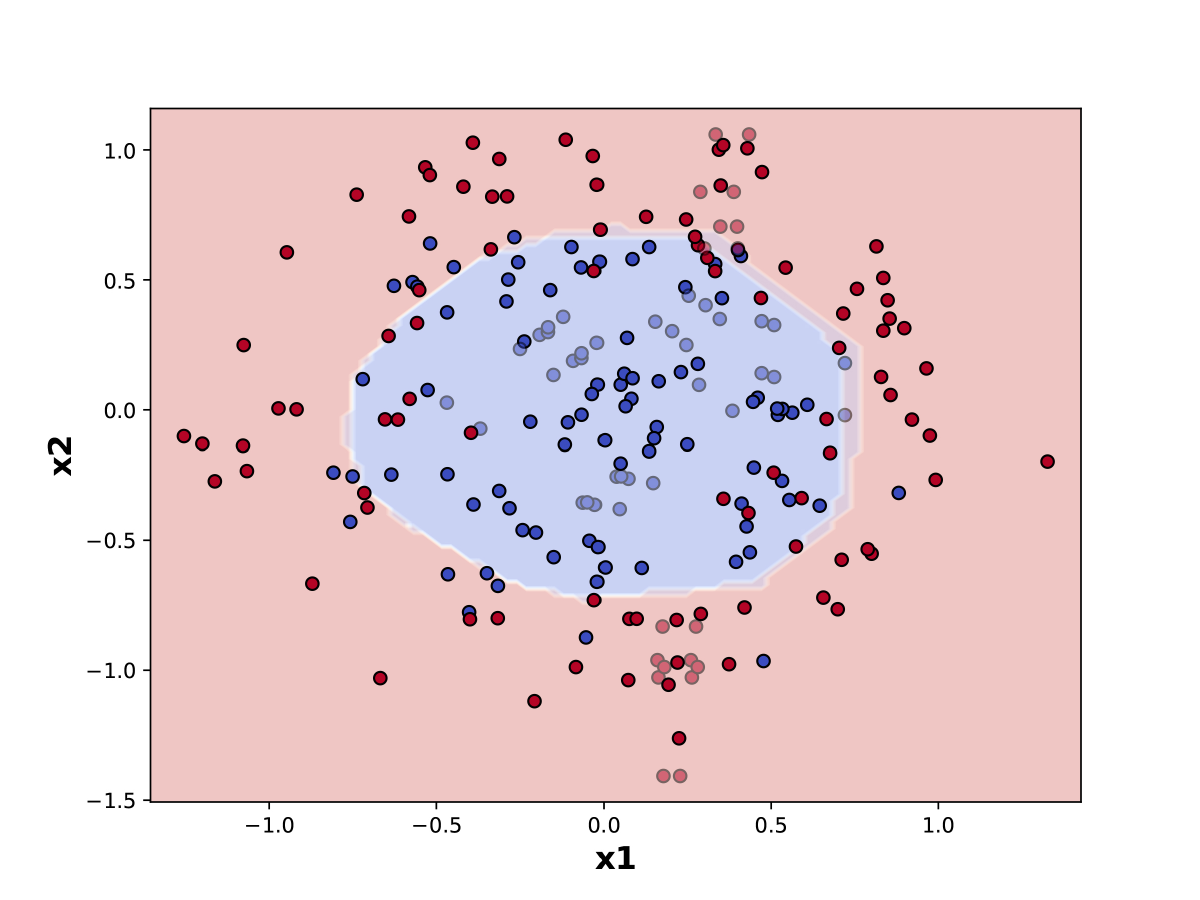}}
    \caption{
    Illustrations on the decision boundaries of PGD-AT and PIAT before/after an epoch. 
    Each subfigure contains the decision boundary illustration before (in light-colored) and after (in dark-colored) the adversarial training.
    % \textcolor{blue}{
    Specifically, the light-colored data represent the 26\(^\text{th}\) epoch of PGD-AT and the 24\(^\text{th}\) epoch of PIAT, while the dark-colored data correspond to the 27\(^\text{th}\) epoch of PGD-AT and the 25\(^\text{th}\) epoch of PIAT.}
    % }
    \label{fig:PIAT decision boundary}
\end{figure*}
We investigate the difference of decision boundary change between PGD-AT and our PIAT framework when the model robustness exhibits a sudden decline.
To facilitate a more comprehensive comparison of the decision boundaries between PGD-AT and PIAT, we enhance the clarity by overlaying the decision boundary images before and after adversarial training.
As illustrated in Fig.~\ref{fig:PIAT decision boundary}, we observe that the decision boundary change of PIAT framework is more moderate than that of PGD-AT when the model robustness exhibits a sudden decrease. 
Taking the 27\(^{th}\) epoch as an example, although the blue data point located in the bottom left near the decision boundary can be classified correctly, some red data points situated in the top right and top left of the decision boundary are misclassified.
On the other hand, taking the 25\(^{th}\) epoch as an example, the decision boundary change in the model trained using the PIAT framework is moderate.
As shown in Fig.~\ref{fig:Accuracy and Robustness}, since the decision boundary change is more moderate, our method effectively enhances the model robustness while maintaining the accuracy of clean examples.

Moreover, we also observe the same phenomenon in the training process on CIFAR10 dataset. 
As shown in Fig.~\ref{fig:adv_classical}, typical advanced adversarial training methods also suffer from apparent oscillations and perform unsatisfactorily on the model robustness.
Compared with these approaches, our method not only alleviates the difficulty of convergence but also improves the performance of model.

Additionally, as illustrated in Fig.~\ref{fig:adv_classical}, our parameter interpolation method can also alleviate the overfitting issue in the later stage of the training.
Theorem~\ref{Theorem 3.1} provides the theoretical analysis of this phenomenon.

\textbf{Theorem 3.1} 
\label{Theorem 3.1}
\emph{Assuming that for \(i,j \in \{1,...,T\}\), \(\boldsymbol{\theta}_i=\boldsymbol{\theta}_j\) if and only if \(i=j\).
Model \(f_{\boldsymbol{\theta}}\) is continuous and at least first-order differentiable.
\(f_{\boldsymbol{\tilde{\theta}}}\) is based on parameter interpolation \(\boldsymbol{\tilde{\theta}}=\lambda \boldsymbol{\theta}_i+(1-\lambda) \boldsymbol{\theta}_{i+1}\).
The difference between the prediction of model \(f_{\boldsymbol{\tilde{\theta}}}\) and model \(f_{\boldsymbol{\theta}_{i}}\) is a first-order infinitesimal of \(\lambda\) if and only if \(\lambda \rightarrow 1\).
}

% The proof of the theorem can be found in Appendix~\ref{Theorem 1}.

\textbf{Proof.} 
For the sake of the first differentiability of \(f_{\boldsymbol{\theta}_{i+1}}(x, y)\), based on the Taylor expansion, we can fit a first order polynomial of \(f_{\boldsymbol{\theta}_{i+1}}(x, y)\) to approximate the value of \(f_{\boldsymbol{\theta}_i}(x, y)\):
\begin{equation}
    f_{\boldsymbol{\theta}_{i+1}}(x, y)=f_{\boldsymbol{\theta}_i}(x, y)+\Delta\boldsymbol{\theta}_1^T\nabla_{\boldsymbol{\theta}_i}f_{\boldsymbol{\theta}_i}(x, y)+O(\Delta\boldsymbol{\theta}_1^n),
\end{equation}
where \(\Delta \boldsymbol{\theta}_1=\boldsymbol{\theta}_{i+1}-\boldsymbol{\theta}_{i}\) and \(O(\Delta\boldsymbol{\theta}_1^n)\) represents the higher order remainder term. Note that the subscript \(\Delta\boldsymbol{\theta}_1\) here stands for a neighborhood where the Taylor expansion approximates a function by polynomials of any point in terms of its value and derivatives.
In the same way, we can get the first order polynomial of \(f_{\boldsymbol{\tilde{\theta}}}(x, y)\) to approximate the value of \(f_{\boldsymbol{\theta_i}}(x, y)\):
\begin{equation}
    f_{\boldsymbol{\tilde{\theta}}}(x, y)=f_{\boldsymbol{\theta}_i}(x, y)+\Delta \boldsymbol{\theta}_2^T\nabla_{\boldsymbol{\theta}_i}f_{\boldsymbol{\theta}_i}(x, y)+O(\Delta\boldsymbol{\theta}_1^n),
\end{equation}
where \(\Delta\boldsymbol{\theta}_2=\boldsymbol{\tilde{\theta}}-\boldsymbol{\theta}_{i}=(1-\lambda)\Delta\boldsymbol{\theta}_1\).
Therefore, the difference between the prediction of model \(f_{\boldsymbol{\tilde{\theta}}}\) and model \(f_{\boldsymbol{\theta}_{i}}\) can be formulated as:

\begin{equation}
\begin{aligned}
    {} f_{\boldsymbol{\tilde{\theta}}}(x, y)-f_{\boldsymbol{\theta_i}}(x, y) & = \Delta\boldsymbol{\theta}_2^T\nabla_{\boldsymbol{\theta}_i}f_{\boldsymbol{\theta}_i}(x, y)+O(\Delta\boldsymbol{\theta}_2^n)\\
    & =(1-\lambda)\Delta\boldsymbol{\theta}_1^T\nabla_{\boldsymbol{\theta}_i}f_{\boldsymbol{\theta}_i}(x, y)+O(\Delta\boldsymbol{\theta}_2^n)\\
    & \leq \Delta\boldsymbol{\theta}_1^T\nabla_{\boldsymbol{\theta}_i}f_{\boldsymbol{\theta}_i}(x, y)+O(\Delta\boldsymbol{\theta}_1^n)\\
    & =f_{\boldsymbol{\theta}_{i+1}}(x, y)-f_{\boldsymbol{\theta}_i}(x, y).
\end{aligned}
\end{equation}

In the later stage, the model trained by standard adversarial training is overfitting. Meanwhile, the hyperparameter \(\lambda\) is close to 1. Thus, the prediction of model \(f_{\boldsymbol{\tilde{\theta}}}\) is more similar to \(f_{\boldsymbol{\theta}_{i}}\) instead of \(f_{\boldsymbol{\theta}_{i+1}}\), alleviating the overfitting issue.

Theorem~3.1 indicates that the difference between the prediction of models \(f_{\boldsymbol{\tilde{\theta}}}\) and \(f_{\boldsymbol{\theta}_{i}}\) is smaller than that of \(f_{\boldsymbol{\theta}_{i+1}}\) and \(f_{\boldsymbol{\theta}_{i}}\) when \(\lambda\) is close to 1.
This reveals the potential reason why the model trained by the parameter interpolation method does not cause the overfitting issue in the later stage.

\subsection{Regularization of Aligning Logits}
\label{Absolute Magnitude vs Relative Magnitude}
Here, we also conduct a similar experiment to study the robust improvement of the regularization.
As shown in Fig.~\ref{fig:Accuracy and Robustness}, although ALP can effectively boost model robustness, the model accuracy decreases apparently at the same time.
Taking the 23\(^{th}\) epoch as an example, the increase of model robustness comes at the sacrifice of accuracy.
We revisit the robustness regularization of ALP, which can be formulated as follows:
\begin{equation}
    \mathcal{L}_{ALP}=\|f_{\boldsymbol{\theta}}(\boldsymbol{x})-f_{\boldsymbol{\theta}}(\boldsymbol{x}^{adv})\|_2^2 ,
\end{equation}
where \(f_{\boldsymbol{\theta}}(\boldsymbol{x})\) is the output logits of the model, and \(||\cdot||_2\) denotes \(l_2\)-norm.
It might be attributed to the fact that clean and adversarial examples belong to different data distributions.
Therefore, simply forcing the output logit to be close is unreasonable, which naturally leads to an opposed relationship between accuracy and robustness.

To maintain the model accuracy while boosting the robustness, we further customize a novel regularization term, which pays more attention to the relative magnitude of logits rather than absolute magnitude.
As shown in Fig.~\ref{fig:Accuracy and Robustness}, the model trained with our proposed regularization term improves model robustness while keeping the clean accuracy.

\section{Methodology}
% 这一节主要介绍了MOMAT算法的实现和与其他对抗训练方法结合。
In this section, we introduce the realization of our Parameter Interpolation Adversarial Training (PIAT) framework and describe how to combine %it with 
our proposed Normalized Mean Square Error (NMSE) regularization term 
to the framework. 

% 介绍MOMAT对抗训练算法的具体实现细节，以及对于参数的分析与选取。
\subsection{The Proposed New Framework: PIAT}
\label{The PIAT Framework}
To mitigate the impact of the rapid changes in the model decision boundary, we propose a new framework called Parameter Interpolation Adversarial Training (PIAT). 
PIAT tunes the model parameters by interpolating the model parameters between the previous and current epochs. 
Mathematically, it can be formalized as follows:
\begin{equation}
    \boldsymbol{\theta}_t'=\lambda\cdot\boldsymbol{\theta}_{t-1}'+(1-\lambda)\cdot\boldsymbol{\theta}_t, \quad 0 \leq \lambda \leq 1 , 
    \label{eq:MOMAT}
\end{equation}
where \(\boldsymbol{\theta}_{t-1}'\) is the model parameters of the previous epoch after interpolation, and \(\boldsymbol{\theta}_t\) is the current parameters at the end of the training epoch before interpolation. 
Before starting the next training epoch, we tune the model parameters to \(\boldsymbol{\theta}_t'\). 
The hyper-parameter \(\lambda\) controls the tradeoff between previous and current parameters. 

Based on the observations presented in Section~\ref{Toy model}, we can gain an intuitive understanding the value of \(\lambda\) from two perspectives. 
Initially, when the model lacks robustness and informative parameters due to insufficient fitting to the training data.
Therefore, \(\lambda\) should be set to a small value in the early stage.
However, as the training progresses and the model becomes more adversarially robust, \(\lambda\) should be gradually increased towards 1 in the later stages of training.

\begin{algorithm}[t]
    \caption{The \name Framework}
    \label{alg: MOMAT}
    \begin{algorithmic}
    \STATE {\bfseries Input:} Initial model parameters $\boldsymbol{\theta}_0$, perturbation step size $\epsilon$, number of adversarial attack steps \(K\), number of epochs $N$, weight function $g(\cdot)$
    % , weighted parameter $\lambda_0$
    \STATE {\bfseries Output:} $\boldsymbol{\theta}_N'$
    \STATE Initialize $\boldsymbol{\theta}_0'\leftarrow \boldsymbol{\theta}_0$
    \FOR{$i=1$ {\bfseries to} $N$}
    \STATE $\boldsymbol{\theta}_i\leftarrow\boldsymbol{\theta}'_{i-1}$
    \FOR{$minibatch$ $\boldsymbol{x}\subset\boldsymbol{x}$}
    \STATE $\boldsymbol{x}^{adv}$ $\leftarrow$ $\boldsymbol{x}$
    \FOR{$k=1$ {\bfseries to} $K$}
    \STATE $\boldsymbol{x}^{adv} \leftarrow\boldsymbol{x}^{adv}+\epsilon \cdot sign(\nabla_{\boldsymbol{x}} \mathcal{L}_{CE}(\boldsymbol{x}^{adv},y))$
    \STATE $\boldsymbol{x}^{adv}\leftarrow clip(\boldsymbol{x}^{adv},\boldsymbol{x}-\epsilon,\boldsymbol{x}+\epsilon)$
    \ENDFOR
    \STATE $loss=\mathcal{L}(\boldsymbol{x}^{adv},y)$
    \STATE update $\boldsymbol{\theta}_i$
    \ENDFOR
    \STATE $\lambda \leftarrow g(i)$ 
    \STATE $\boldsymbol{\theta}_i'\leftarrow\lambda\cdot\boldsymbol{\theta}_{i-1}'+(1-\lambda)\cdot\boldsymbol{\theta}_i$
    \ENDFOR
    \STATE \textbf{return} $\boldsymbol{\theta}_N'$
    \end{algorithmic}
\end{algorithm}
According to the above analysis, \(\lambda\) should change over the course of training instead of using a fixed value. 
The value of \(\lambda\) should be small in the early training stage and gradually increase along with the training, ensuring the convergence speed and alleviating the overfitting issue in the adversarial training process. 
In this paper, we set \(\lambda\) as follows:
\begin{equation}
\label{eq5}
    \lambda=g(n)=\frac{an+b}{cn+d}, \quad  c \geq a, \quad d \geq b ,
\end{equation}
where \(n\) denotes the current number of training epochs. \(a\), \(b\), \(c\) and \(d\) are hyper-parameters and we set \(a = b = c = 1\), \(d = 10\) in this work. 

Algorithm~\ref{alg: MOMAT} summarizes the flexible framework of PIAT, which can seamlessly integrate with different adversarial training methods on both CNNs and ViTs without imposing restrictions on the choice of loss function or model.

\subsection{The Proposed Regularization Term: NMSE}
According to the discussion in Section~\ref{Absolute Magnitude vs Relative Magnitude}, instead of aligning the clean and adversarial examples by classification probabilities, we utilize the output logits normalized with \(l_2\)-norm.

% \textcolor{blue}{
We align the clean and adversarial examples by minimizing the mean square error between their normalized output logits. Besides, we set \((1-p_{clean})\) as the weight for different adversarial examples so that the model will pay more attention to the clean examples that are vulnerable.
% } 
We formulate the Normalized Mean Square Error (NMSE) regularization as follows:
\begin{equation}
    \mathcal{L}_{NMSE}=(1-p_{clean})\cdot
    \left\|\frac{f_{\boldsymbol{\theta}}(\boldsymbol{x})}{||f_{\boldsymbol{\theta}}(\boldsymbol{x})||_{2}}-\frac{f_{\boldsymbol{\theta}}(\boldsymbol{x}^{adv})}{||f_{\boldsymbol{\theta}}(\boldsymbol{x}^{adv})||_{2}}\right\|_2^2,
\end{equation}
where \(\boldsymbol{x}^{adv}\) is the adversarial example, \(f_{\boldsymbol{\theta}}(\boldsymbol{x})\) is the output logits of the model, and \(||\cdot||_2\) denotes \(l_2\)-norm.

\begin{table*}[t]
\caption{The clean and robust accuracy (\%) of our methods (PIAT+NMSE) and defense baselines using ResNet18 model trained on CIFAR10, CIFAR100 and SVHN datasets under various adversarial attacks. We report the results of the best checkpoint according to the highest robust accuracy under PGD20 attack and the final checkpoint. The best result among defense methods in each column is in \textbf{bold}.}
% \linespread{1.25}
\label{table: mitigating all}
\begin{center}
\begin{small}
\resizebox{\textwidth}{!}
{
\begin{tabular}{cccccccccccccc}
\toprule
\multirow{2}{*}{Dataset} & \multirow{2}{*}{Method} &
  \multicolumn{3}{c}{Clean} &
  \multicolumn{3}{c}{PGD20} &
  \multicolumn{3}{c}{CW} &
  \multicolumn{3}{c}{AA}
  \\ \cmidrule(lr){3-5} \cmidrule(lr){6-8} \cmidrule(lr){9-11} \cmidrule(lr){12-14}
&&Best&Final&Diff&Best&Final&Diff&Best&Final&Diff&Best&Final&Diff\\
  \midrule
  \multirow{6}{*}{CIFAR10}
  &PGD-AT   
  &\textbf{84.28}&\textbf{85.62}&~1.34&50.29&45.86&~4.43&49.31&43.25&~6.06&46.33&41.36&~4.97\\
  &ALP
  &79.74&81.45&~1.71&52.37&48.62&~3.75&49.60&43.87&~5.73&46.13&41.88&~4.25\\
  &TRADES 
   & 82.39&83.04&\textbf{~0.65}& 53.60&50.74&~2.86& 50.90&49.04&~1.86&48.04&46.80&~1.24\\
  &MART  
  &81.91&83.99&~2.08&53.70&48.63&~5.07&49.35&44.92&~4.43&47.45&43.65&~3.80\\
  &MAIL
  &82.65&85.17&~2.49&51.15&47.14&~4.01&48.88&44.38&~4.50&45.16&43.02&~2.14\\
  % &Generalist&textbf{89.09}&\textbf{89.53}&\textbf{-0.52}&50.01&45.04&~4.97&50.04&45.62&~4.42&46.07&42.60&~3.47\\
  % &CFA&82.80&83.88&~1.08&53.24&51.69&~1.55&51.45&49.97&~1.48&48.40&47.74&\textbf{~0.64}\\
  %   &RAT&81.63&82.61&~0.98&52.25&50.09&~2.16&49.47&47.93&~1.54&45.20&44.30&~0.90\\
  & \textcolor{black}{CFA} & \textcolor{black}{82.80} & \textcolor{black}{83.88} & \textcolor{black}{~1.08} & \textcolor{black}{53.24} & \textcolor{black}{51.69} & \textcolor{black}{~1.55} & \textcolor{black}{51.45} & \textcolor{black}{49.97} & \textcolor{black}{~1.48} & \textcolor{black}{48.40} & \textcolor{black}{47.74} & \textcolor{black}{\textbf{~0.64}} \\
  & \textcolor{black}{RAT} & \textcolor{black}{81.63} & \textcolor{black}{82.61} & \textcolor{black}{~0.98} & \textcolor{black}{52.25} & \textcolor{black}{50.09} & \textcolor{black}{~2.16} & \textcolor{black}{49.47} & \textcolor{black}{47.93} & \textcolor{black}{~1.54} & \textcolor{black}{45.20} & \textcolor{black}{44.30} & \textcolor{black}{~0.90} \\
  \cmidrule(lr){2-14}
  &\textbf{\name+NMSE}    
  &80.96&82.84&~1.88&\textbf{53.74}&\textbf{52.81}&\textbf{~0.93}&\textbf{51.72}&\textbf{50.49}&\textbf{~1.23}&\textbf{48.80}&\textbf{47.97}&~0.83\\
  \midrule
  \multirow{6}{*}{CIFAR100}
  &PGD-AT   
  & 58.48&58.53&\textbf{~0.05}& 28.36&21.72&~6.64&27.06&21.12&~5.94&23.85&19.55&~4.30\\
  &ALP
  &57.29&58.65&~1.36&28.12&24.66&~3.46&26.84&22.17&~4.67&23.57&20.49&~3.08\\
  &TRADES 
  &56.71&56.32&~0.39& 29.19&27.70&~1.49&26.05&24.53&~1.52&23.91&22.70&~1.21 \\
  &MART 
  &55.26&57.77&~2.51&30.10&25.96&~4.14&26.30&23.79&~2.51&24.13&22.35&~1.78\\
  &MAIL 
  &\textbf{58.73}&59.00&~0.27&27.99&24.69&~3.30&26.28&23.37&~2.91&22.50&20.86&~1.64\\
  % &Generalist&\textbf{61.97}&\textbf{67.25}&~5.28&29.48&21.90&~7.58&27.77&21.31&6.46&23.96&15.61&~8.35\\
    % &CFA&56.28&\textbf{61.62}&~5.34&30.64&29.74&~0.90&27.74&25.95&~1.79&24.26&21.58&~2.68\\
    % &RAT&53.35&56.35&~3.00&28.69&27.67&~1.02&25.52&23.69&~1.83&23.10&22.40&\textbf{~0.70}\\
& \textcolor{black}{CFA} & \textcolor{black}{56.28} & \textcolor{black}{\textbf{61.62}} & \textcolor{black}{~5.34} & \textcolor{black}{30.64} & \textcolor{black}{29.74} & \textcolor{black}{~0.90} & \textcolor{black}{27.74} & \textcolor{black}{25.95} & \textcolor{black}{~1.79} & \textcolor{black}{24.26} & \textcolor{black}{21.58} & \textcolor{black}{~2.68} \\
& \textcolor{black}{RAT} & \textcolor{black}{53.35} & \textcolor{black}{56.35} & \textcolor{black}{~3.00} & \textcolor{black}{28.69} & \textcolor{black}{27.67} & \textcolor{black}{~1.02} & \textcolor{black}{25.52} & \textcolor{black}{23.69} & \textcolor{black}{~1.83} & \textcolor{black}{23.10} & \textcolor{black}{22.40} & \textcolor{black}{\textbf{~0.70}} \\
  \cmidrule(lr){2-14}
  &\textbf{\name+NMSE}    
  &56.04&57.16&~1.12&\textbf{31.45}&\textbf{30.87}&\textbf{~0.58}&\textbf{28.74}&\textbf{27.76}&\textbf{~1.02}&\textbf{26.09}&\textbf{25.13}&~0.96\\
  \midrule
    \multirow{6}{*}{SVHN}
  &PGD-AT  
&\textbf{93.85}&\textbf{94.33}&~0.48& 59.01&52.35&~6.66&48.66&44.13&~4.53&43.02&38.66&~4.36\\
  &ALP &92.54&93.67&~1.13&59.13&55.12&~5.01&52.22&48.53&~3.69&45.67&42.41&~3.26
\\
  &TRADES   & 90.88&91.34&\textbf{~0.46}&59.50&57.04&~2.46&52.76&50.42&~2.34&46.59&44.87&~1.72
\\
  &MART &90.84&92.95&~2.11&57.70&54.29&~3.41&52.95&50.09&~2.86&46.98&43.75&~3.23
\\
  &MAIL &90.15&93.69&~3.54&57.47&54.60&~3.14&52.78&49.73&~3.05&46.26&41.24&~5.02
\\
  % &Generalist&92.88&\textbf{96.83}&-3.95&60.21&42.10&18.11&54.97&43.36&11.61&49.12&29.26&19.86\\
% &CFA&92.23&93.68&~1.45&60.77&58.85&~1.92&55.17&52.71&~2.46&49.64&46.53&~3.11\\
%   &RAT&89.23&90.97&~1.74&45.83&38.25&~7.58&44.15&32.71&11.44&46.10&22.40&23.70\\
 & \textcolor{black}{CFA} & \textcolor{black}{92.23} & \textcolor{black}{93.68} & \textcolor{black}{~1.45} & \textcolor{black}{60.77} & \textcolor{black}{58.85} & \textcolor{black}{~1.92} & \textcolor{black}{55.17} & \textcolor{black}{52.71} & \textcolor{black}{~2.46} & \textcolor{black}{49.64} & \textcolor{black}{46.53} & \textcolor{black}{~3.11} \\
 & \textcolor{black}{RAT} & \textcolor{black}{89.23} & \textcolor{black}{90.97} & \textcolor{black}{~1.74} & \textcolor{black}{45.83} & \textcolor{black}{38.25} & \textcolor{black}{~7.58} & \textcolor{black}{44.15} & \textcolor{black}{32.71} & \textcolor{black}{11.44} & \textcolor{black}{46.10} & \textcolor{black}{22.40} & \textcolor{black}{23.70} \\
  \cmidrule(lr){2-14}
  &\textbf{\name+NMSE}  &91.70&93.07&~1.37& \textbf{61.21}&\textbf{59.84}&\textbf{~1.37}&\textbf{55.88}&\textbf{54.45}&\textbf{~1.43}&\textbf{51.29}&\textbf{49.82}&\textbf{~1.47}
\\
\bottomrule
\end{tabular}
}
\end{small}
\end{center}
\centering
\end{table*}
\begin{table}[t]
\caption{The clean and robust accuracy (\%) of our methods (PIAT+NMSE) and defense baselines using WRN-32-10 model on CIFAR10 and CIFAR100 datasets. 
The best result in each column is in \textbf{bold}.
}
\label{tab:PIAT_WRN}
\begin{center}
\begin{small}
% \begin{sc}
\resizebox{\columnwidth}{!}
{
\begin{tabular}{ccccc}
\toprule
Dataset& Method &Clean& PGD20& AA\\
\midrule
\multirow{6}{*}{CIFAR10} 
  &PGD-AT  & \textbf{86.87}&48.77&47.78\\
  &ALP &84.18&53.55&49.68\\
  &TRADES & 82.13&55.14&50.38\\
  &MART&81.57&56.44&49.58\\
  &MAIL&84.96&52.58&47.26\\
  % &Generalist&\textbf{91.03}&56.88&52.91\\
  % &CFA&86.44&57.84&52.96\\
  %   &RAT&83.46&57.07&51.56\\
  & \textcolor{black}{CFA} & \textcolor{black}{86.44} & \textcolor{black}{57.84} & \textcolor{black}{52.96} \\
  & \textcolor{black}{RAT} & \textcolor{black}{83.46} & \textcolor{black}{57.07} & \textcolor{black}{51.56} \\
\cmidrule(lr){2-5}
  &\textbf{\name+NMSE}&85.04&\textbf{58.04}&\textbf{53.83}\\
  \midrule
  \multirow{6}{*}{CIFAR100}
  &PGD-AT&59.30&28.13&23.99\\
  &ALP &58.11&28.59&24.45\\
  &TRADES&57.99&31.97&26.76\\
  &MART&55.19&31.16&26.46\\
  &MAIL&58.04&29.50&23.97\\
  % &Generalist&\textbf{64.62}&33.77&28.94\\
  % &CFA&\textbf{63.37}&33.89&28.98\\
  %   &RAT&60.89&33.39&27.95\\
  & \textcolor{black}{CFA} & \textcolor{black}{\textbf{63.37}} & \textcolor{black}{33.89} & \textcolor{black}{28.98} \\
  & \textcolor{black}{RAT} & \textcolor{black}{60.89} & \textcolor{black}{33.39} & \textcolor{black}{27.95} \\
\cmidrule(lr){2-5}
  &\textbf{\name+NMSE}& 61.04&\textbf{35.15}&\textbf{30.07}\\
  \bottomrule
\end{tabular}
}
\end{small}
\end{center}
\centering
\end{table}
\begin{figure*}
    \centering
    \subfloat[CIFAR10]{
    \includegraphics[width=0.32\textwidth]{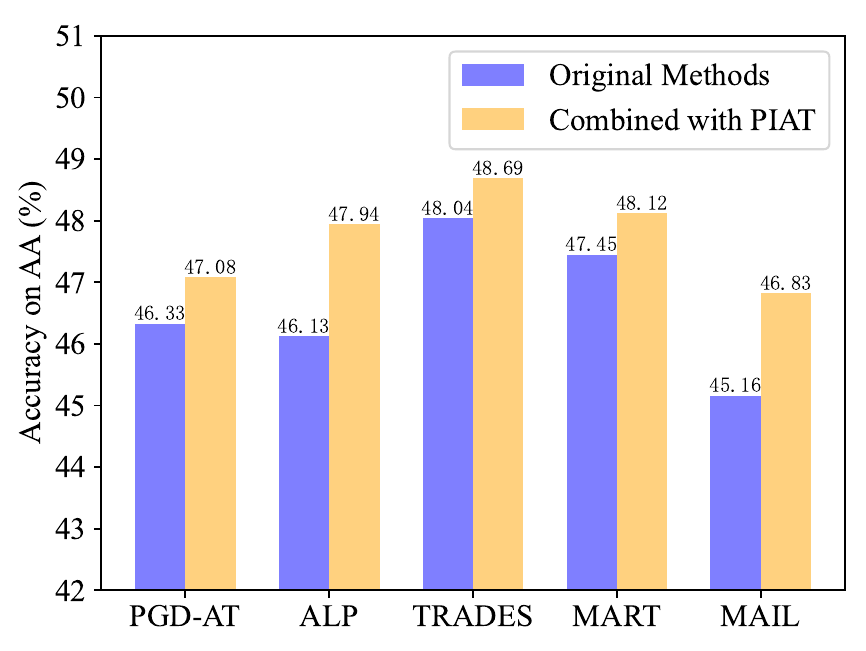}}
    \subfloat[CIFAR100]{
    \includegraphics[width=0.32\textwidth]{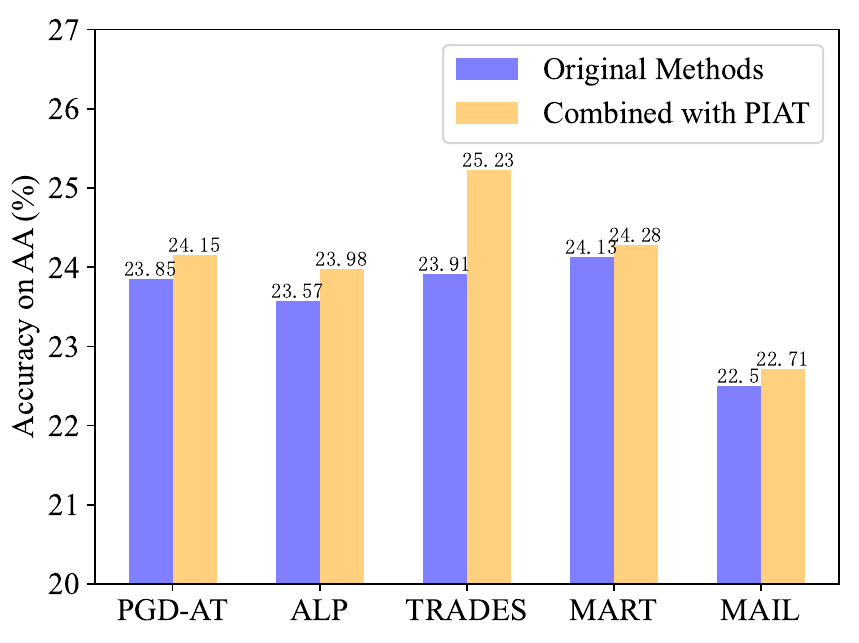}}
    \subfloat[SVHN]
    {\includegraphics[width=0.32\textwidth]{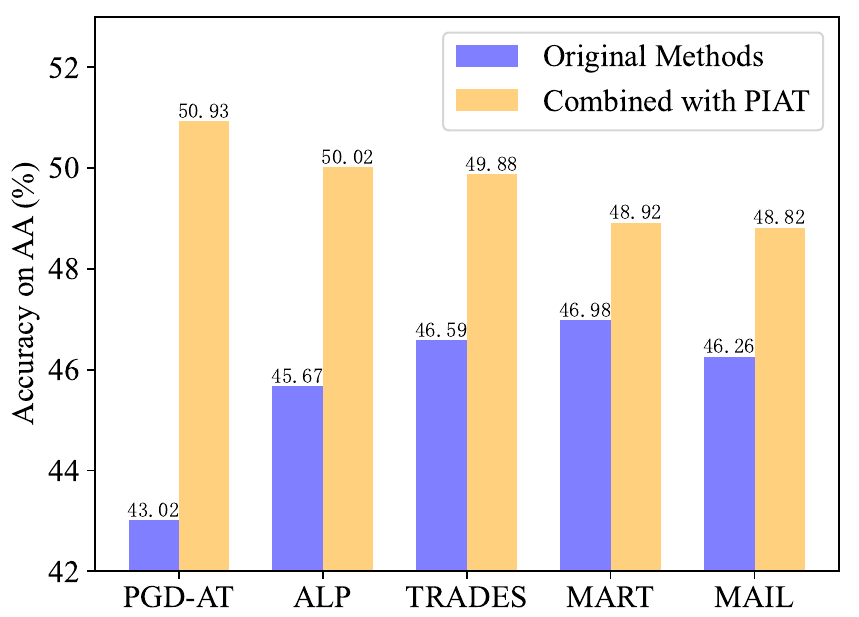}}
    \caption{The robust accuracy (\%) of the \name framework combined with various adversarial training baseline methods under the AA attack on CIFAR10, CIFAR100, and SVHN datasets using ResNet18 model.
    }
    % \vspace{-1.0em}
    \label{fig:MOMAT combined with classical adversarial training}
\end{figure*}

\begin{table}[t]
\caption{The clean and robust accuracy (\%) of \name framework combined with various adversarial training methods on CIFAR10 and CIFAR100 dataset using WRN-32-10 model. The best result among defense methods in each column is in \textbf{bold}.}
\label{tab:PIAT_WRN_AT}
\begin{center}
\begin{small}
\resizebox{\columnwidth}{!}
{
\begin{tabular}{ccccc}
\toprule
Dataset& Method &Clean& PGD20& AA\\
\midrule
\multirow{10}{*}{CIFAR10} 
  &PGD-AT & \textbf{86.87}&48.77&47.78   \\
  &\textbf{\name}& 85.56&\textbf{52.80}&\textbf{48.35}\\
  \cmidrule(lr){2-5} 
  &ALP &\textbf{84.18}&53.55&49.68\\
  &\textbf{\name}+ALP&83.35&\textbf{57.71}&\textbf{52.35}\\
    \cmidrule(lr){2-5}
  &TRADES & \textbf{82.13}&55.14&50.38\\
  &\textbf{\name}+TRADES&82.08&\textbf{58.93}&\textbf{53.73}\\
  \cmidrule(lr){2-5} 
  &MART&\textbf{81.57}&56.44&49.58\\
  &\textbf{\name}+MART&79.88&\textbf{59.51}&\textbf{52.84}\\
  \cmidrule(lr){2-5} 
  &MAIL&\textbf{84.96}&52.58&47.26\\
  &\textbf{\name}+MAIL&84.24&\textbf{57.53}&\textbf{52.22}\\
  \midrule
  \multirow{10}{*}{CIFAR100} 
  &PGD-AT&59.30&28.13&23.99   \\
  &\textbf{\name}&\textbf{60.09}&\textbf{34.46}&\textbf{29.47}\\
  \cmidrule(lr){2-5} 
  &ALP &58.11&28.59&24.45\\
  &\textbf{\name}+ALP&\textbf{59.25}&\textbf{34.04}&\textbf{28.94}\\
  \cmidrule(lr){2-5} 
  &TRADES&57.99&31.97&26.76\\
  &\textbf{\name}+TRADES& \textbf{59.78}&\textbf{34.52}&\textbf{29.25}\\
  \cmidrule(lr){2-5} 
  &MART&\textbf{55.19}&31.16&26.46\\
  &\textbf{\name}+MART& 54.32&\textbf{34.87}&\textbf{28.79}\\
  \cmidrule(lr){2-5} 
  &MAIL&58.04&29.50&23.97\\
  &\textbf{\name}+MAIL&\textbf{58.52}&\textbf{33.65}&\textbf{28.00}\\
  \bottomrule
\end{tabular}
}
\end{small}
\end{center}
\centering
\end{table}

In summary, the overall loss function in our PIAT framework with NMSE is as follows:
\begin{equation}
    \mathcal{L}= \mathcal{L}_{CE}+ \mu \cdot \mathcal{L}_{NMSE}, 
    \label{Total Loss}
\end{equation}
where \(\mu\) is a hyper-parameter to trade off the cross entropy loss \(\mathcal{L}_{CE}\) on adversarial examples and the NMSE regularization term \(\mathcal{L}_{NMSE}\).
% \textcolor{blue}{
Moreover, we could replace the loss function in PIAT framework to combine with various AT methods.
% }

\section{Experiments}
\subsection{Experimental Setup}
\textbf{Datasets and Models.} 
Following the setting on Generalist~\cite{Generalist}, we conduct experiments on three benchmark datasets including CIFAR10, CIFAR100~\cite{CIFAR10}, and SVHN~\cite{SVHN} under \(L_\infty\) norm. 
The CIFAR10 contains 60000 color images with the size of 32 \(\times\) 32 in 10 classes. 
The CIFAR100 shares the same setting as CIFAR10, except it owns 100 classes consisting of 600 images each.
In CIFAR10 and CIFAR100 datasets, 50000 images are for training, and 10000 images are for testing the performance.
SVHN is a dataset of street view house numbers, which includes 73257 examples for training and 26032 examples for evaluation.
All images are normalized into \([0, 1]\). 
We do the evaluation on two CNNs and three ViTs, including ResNet18~\cite{resnet}, Wide-ResNet-32-10 (WRN-32-10)~\cite{WRN} and ViT~\cite{ViT}, DeiT~\cite{DeiT}, ConViT~\cite{ConViT}, to verify the efficacy of our method.

\textbf{Training and Evaluation Settings.}  
For all the experiments of CNNs, we train ResNet18 (WRN-32-10) using SGD with a momentum of 0.9 for 120 (180) epochs. 
The weight decay is $3.5\times10^{-3}$ for ResNet18 and $7\times 10^{-4}$ for WRN-32-10 on the three datasets.
The initial learning rate for ResNet18 (WRN-32-10) is 0.01 (0.1) till epoch 60 (90) and then linearly decays to 0.001 (0.01), 0.0001 (0.001) at epoch 90 (135) and 120 (180). 
We adopt PGD attack with 10 steps for adversary generation during the training stage. The maximum adversarial perturbation of each pixel is $\epsilon=8/255$ with the step size $\alpha=2/255$. For the TRADES baseline, we adopt $\beta=6$ for the best robustness. For the NMSE regularization term, we set \(\mu=5\) to achieve the best performance.

For the experiments of ViT (ConViT-Base, ViT-Base, DeiT-Small), we follow the previous setting~\cite{ARDPRM} to finetune the various ViTs.
Specifically, models are pre-trained on ImageNet-1K and are adversarially trained for 40 epochs using SGD with weight decay \(1\times10^{-4}\), and an initial learning rate of 0.1 that is divided by 10 at the 36\(^{th}\) and 38\(^{th}\) epochs.
Simple data augmentations such as random crop with padding and random horizontal flips are applied.

We compare the PIAT integrated with NMSE regularization with the following AT baselines: ALP~\cite{ALP}, TRADES~\cite{TRADES}, MART~\cite{MART}, MAIL~\cite{MAIL}, 
% \textcolor{blue}{
CFA~\cite{CFA}, RAT~\cite{RAT}
% } 
on CNNs.
Moreover, we also evaluate the performance of PIAT integrated with ARD and PRM (A\&P)~\cite{ARDPRM} on ViTs. 
We adopt various adversarial attacks to evaluate the defense efficacy of our method, including PGD~\cite{PGDAT}, CW~\cite{CW} and AutoAttack (AA)~\cite{AA}.

\begin{table}[t]
\caption{The clean and robust accuracy (\%) of \name framework combined with other adversarial training methods using ViTs on CIFAR10 dataset. Note that `B' denotes `base', `S' denotes `small'. The best result in each column is highlighted in \textbf{bold}.}
\label{tab:PIAT_ViT}
\begin{center}
\begin{small}
\resizebox{\columnwidth}{!}
{
\begin{tabular}{ccccc}
\toprule
Model& Method &Clean& PGD20& AA\\
\midrule
\multirow{9}{*}{ConViT-B}
  &PGD-AT & 61.47&38.64&34.07\\
  &A\&P &85.21&53.25&49.01\\
  &\textbf{PIAT}+A\&P &\textbf{87.50}&\textbf{56.25}&\textbf{51.86}\\
  \cmidrule(lr){2-5}
    &TRADES& 82.75&52.77&49.61\\
  &A\&P&83.51&53.21&50.11\\
  &\textbf{PIAT}+A\&P&\textbf{86.03}&\textbf{55.88}&\textbf{52.35}\\
  \cmidrule(lr){2-5}
    &MART  &63.61&42.51&37.08\\
  &A\&P&80.32&53.11&48.35\\
  &{PIAT}+A\&P&\textbf{81.20}&\textbf{56.17}&\textbf{51.13}\\
\midrule
\multirow{9}{*}{ViT-B}
  &PGD-AT  & 83.07&52.93&48.99\\
  &A\&P&84.64&53.44&49.67\\
  &\textbf{PIAT}+A\&P&\textbf{87.83}&\textbf{56.34}&\textbf{52.27}\\
  \cmidrule(lr){2-5}
    &TRADES  &83.45&53.07&49.76\\
  &A\&P&83.91&53.52&50.56\\
  &\textbf{PIAT}+A\&P&\textbf{87.09}&\textbf{55.97}&\textbf{52.57}\\
  \cmidrule(lr){2-5}
    &MART&77.05&52.99&47.95\\
  &A\&P&78.75&53.51&49.01\\
  &\textbf{PIAT}+A\&P&\textbf{82.23}&\textbf{55.94}&\textbf{51.03}\\
\midrule
\multirow{9}{*}{DeiT-S}
  &PGD-AT  & 81.36&47.94&47.28\\
  &A\&P &83.08&52.28&47.92\\
  &\textbf{PIAT}+A\&P&\textbf{83.22}&\textbf{53.55}&\textbf{49.31}\\
  \cmidrule(lr){2-5}
    &TRADES &82.32&52.52&49.07\\
  &A\&P&82.81&52.74&49.40\\
  &\textbf{PIAT}+A\&P&\textbf{83.43}&\textbf{53.31}&\textbf{50.10}\\
  \cmidrule(lr){2-5}
    &MART & 76.77&52.06&47.06\\
  &A\&P&\textbf{78.43}&52.98&47.94\\
  &\textbf{PIAT}+A\&P&78.39&\textbf{53.83}&\textbf{48.97}\\
  \bottomrule
\end{tabular}
}
\end{small}
\end{center}
\centering
\end{table}

\subsection{Evaluation on Defense Efficacy}
\label{Result}
We compare the defense efficacy of our method with four AT baselines including PGD-AT, TRADES, MART, MAIL. 
Table~\ref{table: mitigating all} reports the best and final clean and robust accuracy of the ResNet18 model trained using our method or the defense baselines under various adversarial attacks on three datasets.

As shown in Table~\ref{table: mitigating all}, our method exhibits the best robustness on all three datasets 
% \textcolor{blue}{
under PGD20, CW and AA attacks
% }
.
Specifically, our method achieves 48.80\%, 26.09\% and 51.29\% accuracy under the AA attack, surpassing the best results of other defense baselines by 0.40\%, 1.96\%, 2.17\%, respectively.
Notably, the exceptional performance of our method on CIFAR100 highlights its generalizability in handling more complex datasets with a greater number of classes. 

Moreover, our experimental results actually verify Theorem~\ref{Theorem 3.1}, which reveals that our method can mitigate overfitting issues in adversarial training. 
As shown in Table~\ref{table: mitigating all}, our method achieves superior robust accuracy compared to the defense baselines on both the best and the final checkpoints with a minimal gap in robustness between them, indicating that the robustness of the model remains stable during training and the overfitting issue is alleviated.

To further investigate the effectiveness of our method with different network architectures, we conduct similar experiments using the WRN-32-10 model. 
As depicted in Table~\ref{tab:PIAT_WRN}, the results indicate that our method still outperforms the competitors under the PGD and AA attacks, confirming its effectiveness even as the size of the DNN model scales up.

\begin{table}[t]
\caption{The clean and robust accuracy (\%) of NMSE and ALP under adversarial attacks on CIFAR10 and CIFAR100 datasets with ResNet18 model. The best result among defense methods in each column is in \textbf{bold}.}
\label{tab: ALP_NMSE}
\centering
\begin{center}
\begin{small}
% \begin{sc}
\begin{tabular}{cccccccc}
\toprule
Dataset & Method &Clean & PGD20 & AA \\
\midrule
\multirow{2}*{CIFAR10}
&ALP&79.74&\textbf{52.37}&46.13\\
&\textbf{NMSE}&\textbf{84.77}&51.56&\textbf{46.60}\\
\midrule
\multirow{2}*{CIFAR100}
&ALP&57.29&28.12&23.57\\
&\textbf{NMSE}&\textbf{58.88}&\textbf{29.55}&\textbf{24.82}\\
\bottomrule
\end{tabular}
% \end{sc}
\end{small}
\end{center}
\end{table}

\subsection{Ablation Study}
\label{Ablation Study}
\textbf{PIAT Framework.} 
Since PIAT is a general framework, we incorporate other adversarial training methods into PIAT to demonstrate its defense efficacy. 
Specifically, we evaluate the robust accuracy of the PIAT framework combined with PGD-AT, ALP, TRADES, MART, and MAIL under the AA attack on three datasets, respectively. 
As shown in Fig.~\ref{fig:MOMAT combined with classical adversarial training}, PIAT boosts the robustness of various adversarial training methods against the AA attack over all the three datasets with ResNet18 model. 
The results demonstrate that we can easily incorporate other adversarial training methods into our PIAT framework without incurring any additional cost to achieve better performance.

We also conduct similar experiments on the WRN-32-10 model and three different ViTs. We report the results in Table~\ref{tab:PIAT_WRN_AT} and Table~\ref{tab:PIAT_ViT}, respectively. 
Our PIAT framework integrated with other adversarial training methods significantly enhances the robust accuracy while maintaining the clean accuracy.
For the WRN-32-10 model, when integrated with PIAT, the original adversarial training methods gain an improvement of 0.57\%, 2.67\%, 3.35\%, 3.26\% and 4.96\%, respectively, under AA attack.
Similarly, the combination of A\&P and PIAT significantly enhances the robustness of ViTs, gaining an improvement of 2.85\%, 2.24\%, and 2.78\%, respectively, against AA attacks for ConViT-B.
Our framework leads to higher robust accuracy when combined with other adversarial training methods on both CNNs and ViTs, indicating that PIAT has good flexibility and generalization.

To evaluate the effectiveness of our PIAT framework on different datasets, we also compare the defense performance when combined PIAT with other baseline methods.
As shown in Table~\ref{tab:PIAT_WRN_AT}, our PIAT framework demonstrates a significantly enhancement in the robustness of the model.
Specifically, when combined with the baseline methods, the original baseline methods gain an improvement of 5.48\%, 4.49\%, 2.49\%, 2.33\%, and 4.03\% under AA attack, respectively. 
The results indicate that our PIAT framework is general and effective on different DNNs.

\textbf{NMSE Regularization.} 
To evaluate the effectiveness of our proposed NMSE regularization, we compare the performance of PGD-AT with ALP~\cite{ALP} and NMSE regularization, respectively. 
Table~\ref{tab: ALP_NMSE} presents the accuracy of the ResNet18 model against PGD and AA attacks.
Specifically, the experimental results show that NMSE regularization achieves an absolute improvement of 0.47\% and 1.25\% under AA attack on CIFAR-10 and CIFAR-100, respectively. 
The experimental results demonstrate that our NMSE regularization surpasses ALP in both clean accuracy and robust accuracy.
Moreover, the improvements achieved by NMSE regularization are significant, highlighting its effectiveness in enhancing model robustness against strong adversarial attacks. 
The superior performance indicates the potential of NMSE regularization as a reliable method for improving the adversarial robustness of DNNs.

\subsection{Further Study}
\begin{figure}[t]
% \begin{minipage}{0.35\linewidth}
    \centering
    {
    \includegraphics[width=0.45\textwidth]{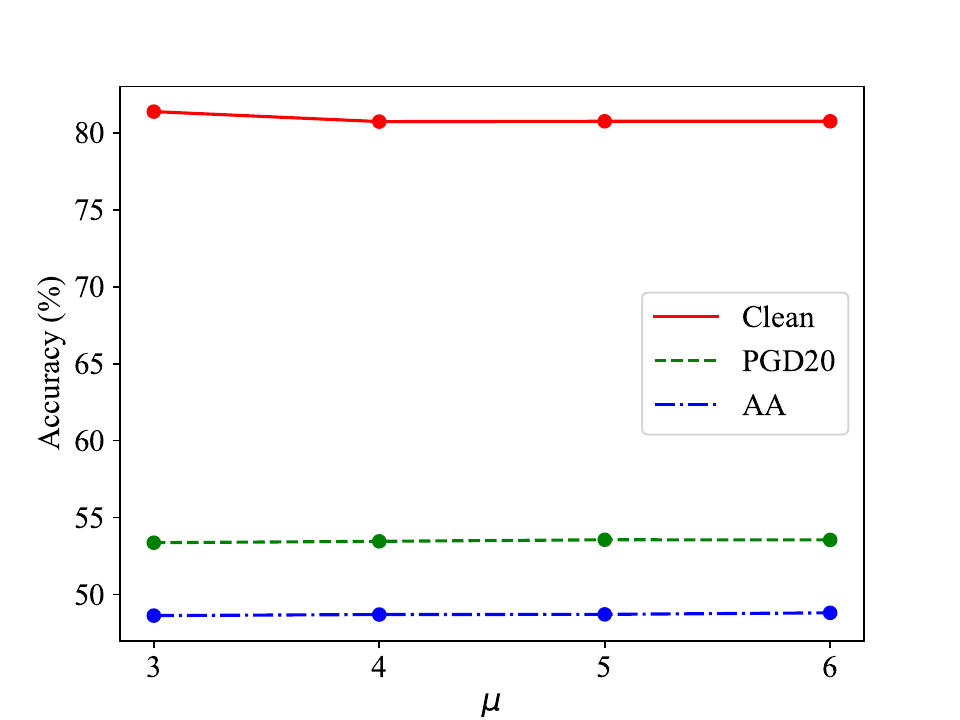}}
    \caption{The clean and robust accuracy (\%) on CIFAR10 dataset for different hyper-parameters of the NMSE regularization term when combined with the \name framework. The performance of NMSE indicates the robustness of the hyper-parameter \(\mu\).}
    \label{fig:Hyperparameter of NMSE}
% \end{minipage}
% ~~~~~~~~~~
% \begin{minipage}{0.6\linewidth}
% \centering
% % \caption{table}{The clean and robust accuracy (\%) of \name with other advanced adversarial training methods on CIFAR100 dataset with WRN-32-10 model.}
% \label{tab:PIAT_WRN_AT_appendix}
% \captionof{table}{The clean and robust accuracy (\%) of \name with other advanced adversarial training methods on CIFAR100 dataset with WRN-32-10 model.}
% \begin{center}
% \begin{small}
% \begin{tabular}{cccccc}
% \toprule
% Method&Clean& PGD20& PGD100&CW&AA\\
% \midrule
% \multirow{10}{*}{}  
%   PGD-AT&59.30&28.13&28.06&27.01&23.99   \\
%   \textbf{\name}&\textbf{60.09}&\textbf{34.46}&\textbf{34.41}&32.44&\textbf{29.47}\\
%   \cmidrule(lr){1-6} 
%   ALP &58.11&28.59&28.57&27.04&24.45\\
%   \textbf{\name}+ALP&\textbf{59.25}&\textbf{34.04}&\textbf{34.03}&\textbf{31.54}&\textbf{28.94}\\
%   \cmidrule(lr){1-6}  
%   TRADES&57.99&31.97&31.85&27.27&26.76\\
%   \textbf{\name}+TRADES& \textbf{59.78}&\textbf{34.52}&\textbf{34.48}&\textbf{30.69}&\textbf{29.25}\\
%   \cmidrule(lr){1-6} 
%   MART&\textbf{55.19}&31.16&31.02&29.98&26.46\\
%   \textbf{\name}+MART& 54.32&\textbf{34.87}&\textbf{34.94}&\textbf{30.74}&\textbf{28.79}\\
%   \cmidrule(lr){1-6} 
%   MAIL&58.04&29.50&29.45&27.29&23.97\\
%   \textbf{\name}+MAIL&\textbf{58.52}&\textbf{33.65}&\textbf{33.65}&\textbf{30.99}&\textbf{28.00}\\
%   \bottomrule
% \end{tabular}
% \label{tab:PIAT_WRN_AT_appendix}
% \end{small}
% \end{center}
% \centering
% \end{minipage}  
\end{figure}

\textbf{Hyper-parameter of NMSE.}
The hyper-parameter \(\mu\) in Eq.~\ref{Total Loss} trades off the cross-entropy loss and the NMSE regularization term for adversarial examples. 
To investigate the impact of different \(\mu\) values on the accuracy of the PIAT framework combined with NMSE, we conduct a series of experiments on the CIFAR10 dataset.
Fig.~\ref{fig:Hyperparameter of NMSE} presents the results on the CIFAR10 dataset when we take \(\mu=3,4,5,6\).
It demonstrates that the defense effectiveness of our method remains relatively stable across different \(\mu\) values.
This stability indicates that our NMSE regularization term is robust to variations in the \(\mu\) hyper-parameter.
Given these observations, we choose \(\mu=5\) for our experiments, as this value provides an optimal balance between maintaining accuracy on clean examples and robustness against adversarial attacks. The robustness of our method to different \(\mu\) values highlights the effectiveness and reliability of the NMSE regularization term with the PIAT framework.

\begin{figure*}[t]
    \centering
    \begin{minipage}[t]{0.45\textwidth}
        \centering
        \subfloat[Loss landscape of PGD-AT]{        \includegraphics[width=0.9\textwidth]{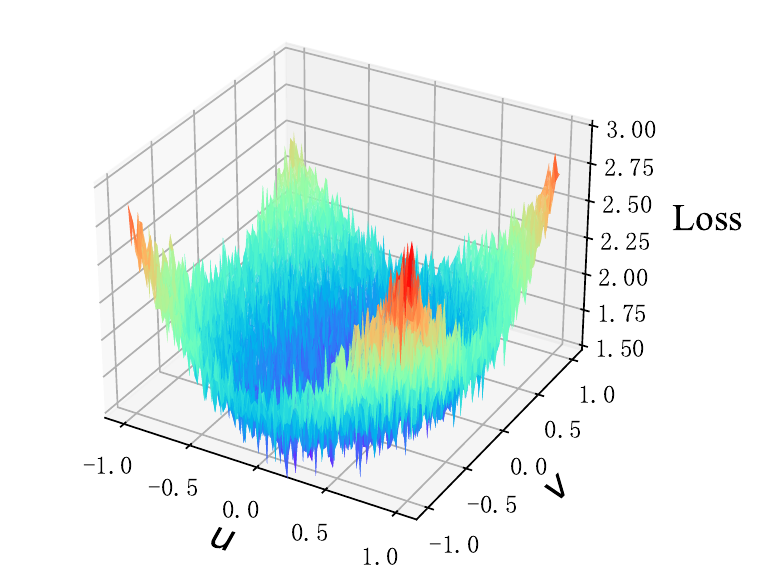}}
    \end{minipage}%
    \begin{minipage}[t]{0.45\textwidth}
        \centering
        \subfloat[Loss landscape of \name]{
        \includegraphics[width=0.9\textwidth]{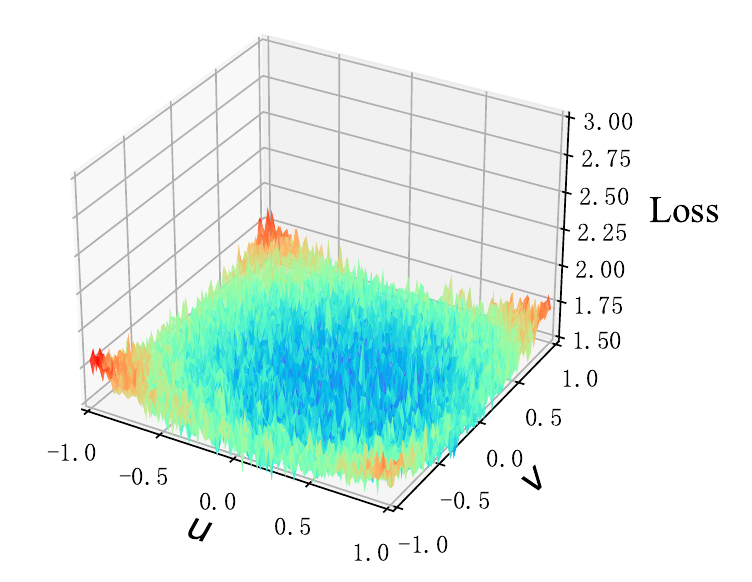}}
    \end{minipage}
    \caption{Illustrations of the loss landscape of PGD-AT and \name in 3D. The loss landscape of standard adversarially trained model changes dramatically, while our \name's changes smoothly, indicating that the model parameters trained by \name converge better to the flatter region.
    }
    \label{fig:MOMAT_model_loss}
\end{figure*}
\begin{figure}[t]
        \centering
        \includegraphics[width=0.4\textwidth]{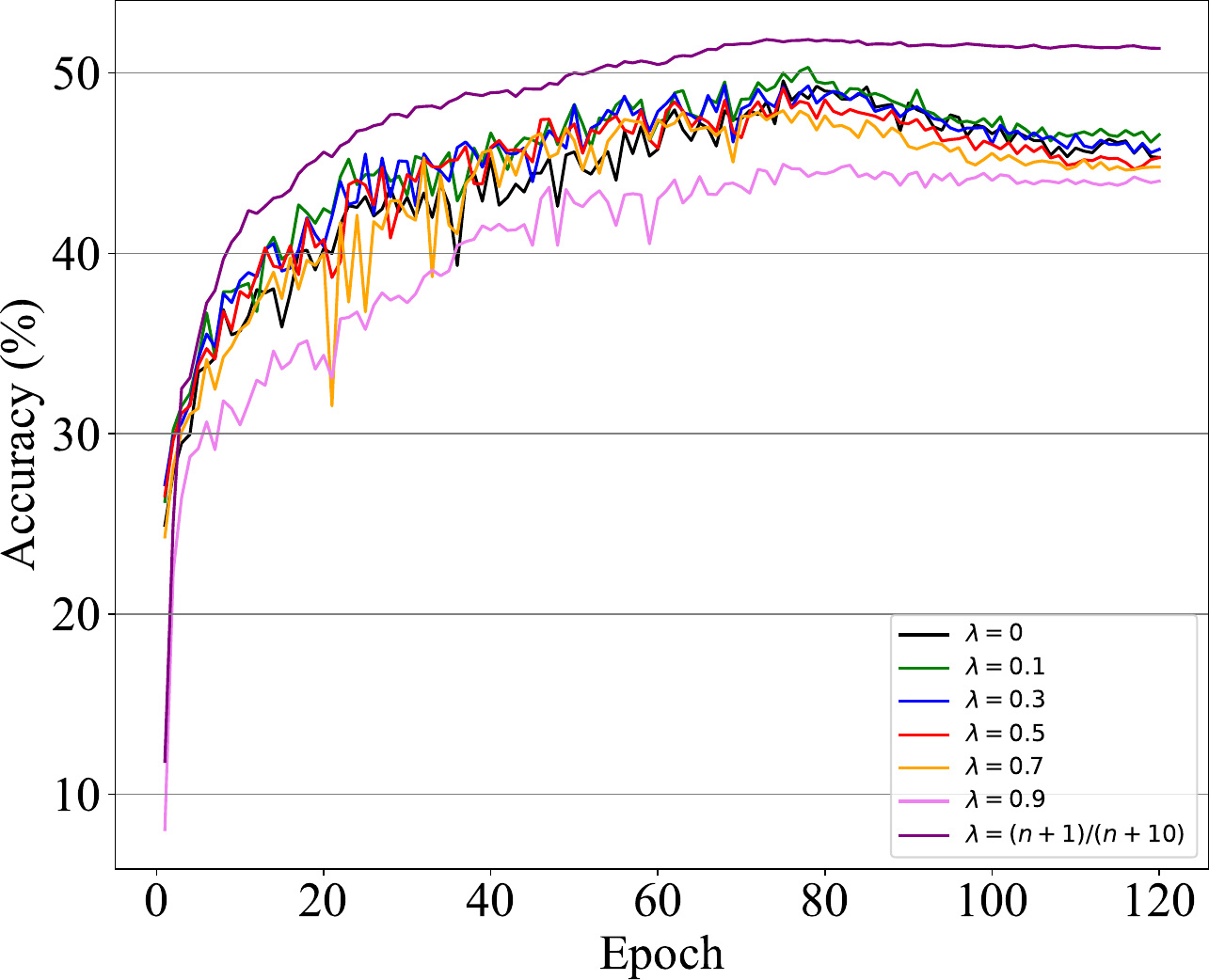}
    \caption{
    The robust accuracy (\%) on adversarial examples of ResNet18 model trained by \name with different \(\lambda\) on the CIFAR10 dataset. \(n\) denotes the current number of training epochs.
    }
    \label{fig:MOMAT_test_accuracy}
\end{figure}

\textbf{Hyper-parameter of PIAT.}
The hyper-parameter \(\lambda\) in Eq.~\ref{eq:MOMAT} controls the trade-off between model parameters from the previous and current epochs. 
In Section~\ref{The PIAT Framework}, we propose dynamically adjusting \(\lambda\) throughout the training instead of using a fixed value. 
To validate our assumption, we evaluate the clean and robust accuracy of PIAT combined with NMSE under PGD20 attack, comparing fixed \(\lambda\) values to our variable \(\lambda\) as defined in Eq.~\ref{eq5}. 
As illustrated in Fig.~\ref{fig:MOMAT_test_accuracy}, during the early stage of training, using a small fixed \(\lambda\) exhibits better model robustness and efficiency compared to a large fixed \(\lambda\).
However, in the later stage, the interpolation with a large fixed \(\lambda\) does not exhibit overfitting issues, which differs from the small fixed \(\lambda\).
These observations indicate that appropriately adjusting \(\lambda\) crucial. 
A dynamic \(\lambda\) alleviates oscillations in the early stage and address the overfitting issues in the later stages of the adversarial training process. 
Thus, the dynamic adjustment plays a key role in enhancing the overall effectiveness and robustness of the model.

\subsection{Loss Landscape}
To provide a comprehensive evaluation of the efficacy of our PIAT framework, we compare the loss landscape of models trained using the PIAT framework and PGD-AT in 3D. 
Let $\boldsymbol{u}$ and $\boldsymbol{v}$ be two random direction vectors sampled from the Gaussian distribution. 
We plot the loss landscape around $\boldsymbol{\theta}$ using the following equation while inputting the same data, where \(m_1,m_2 \in [-1,1]\):
\begin{equation}
    \mathcal{L}(\boldsymbol{\theta};\boldsymbol{u};\boldsymbol{v})=\mathcal{L}\left(\boldsymbol{\theta}+m_1\frac{\boldsymbol{u}}{||\boldsymbol{u}||}+m_2\frac{\boldsymbol{v}}{||\boldsymbol{v|}|}\right).
\end{equation}

As illustrated in Fig.~\ref{fig:MOMAT_model_loss}, we observe that compared with PGD-AT, the model trained using the PIAT framework exhibits less fluctuation in the loss landscape with the changes in model parameters 
% \textcolor{blue}{
under PGD20 attack
% }
.
Furthermore, in comparison to the landscape obtained using PGD-AT, the landscape resulting from the PIAT framework suggests that the model converges to a flatter region.
The flatter region signifies a higher level of robust accuracy, indicating that our PIAT framework improves model stability against adversarial perturbations.
The stability implies that the PIAT framework not only improves robustness but also contributes to better generalization and resilience of the model.
% \textcolor{blue}{
Additionally, the loss landscape of PGD-AT and PIAT framework have similar pattern in other adversarial attacks.
% }

\section{Conclusion}
To mitigate the oscillation and overfitting issues during the training process, we proposed a novel adversarial training framework called PIAT, which interpolates parameter interpolation between previous and current epochs.
Furthermore, we suggested using Normalized Mean Squared Error (NMSE) as a regularization term to align the output of clean and adversarial examples. 
NMSE focuses on the relative magnitude of the logits rather than the absolute magnitude. 
Extensive experiments conducted on multiple benchmark datasets demonstrate the effectiveness of our framework in enhancing the robustness of both Convolutional Neural Networks (CNNs) and Vision Transformers (ViTs).
Moreover, PIAT is flexible and versatile, allowing for the integration of various adversarial training methods into our framework to further boost the performance.

% \textcolor{blue}{
Compared to other methods, our approach can further enhance the adversarial robustness of the model without changing the existing adversarial training framework. We hope that future research will lead to more universal adversarial training frameworks that can further improve classification accuracy as well as robustness of the models.
% }

% \section*{Acknowledgments}
This work is supported by National Natural Science Foundation of China (U22B2017, 62076105).

% \section*{Acknowledgments}
% This should be a simple paragraph before the References to thank those individuals and institutions who have supported your work on this article.

\bibliographystyle{IEEEtran}
\bibliography{main}

\begin{IEEEbiography}[{\includegraphics[width=1in,height=1.25in,clip,keepaspectratio]{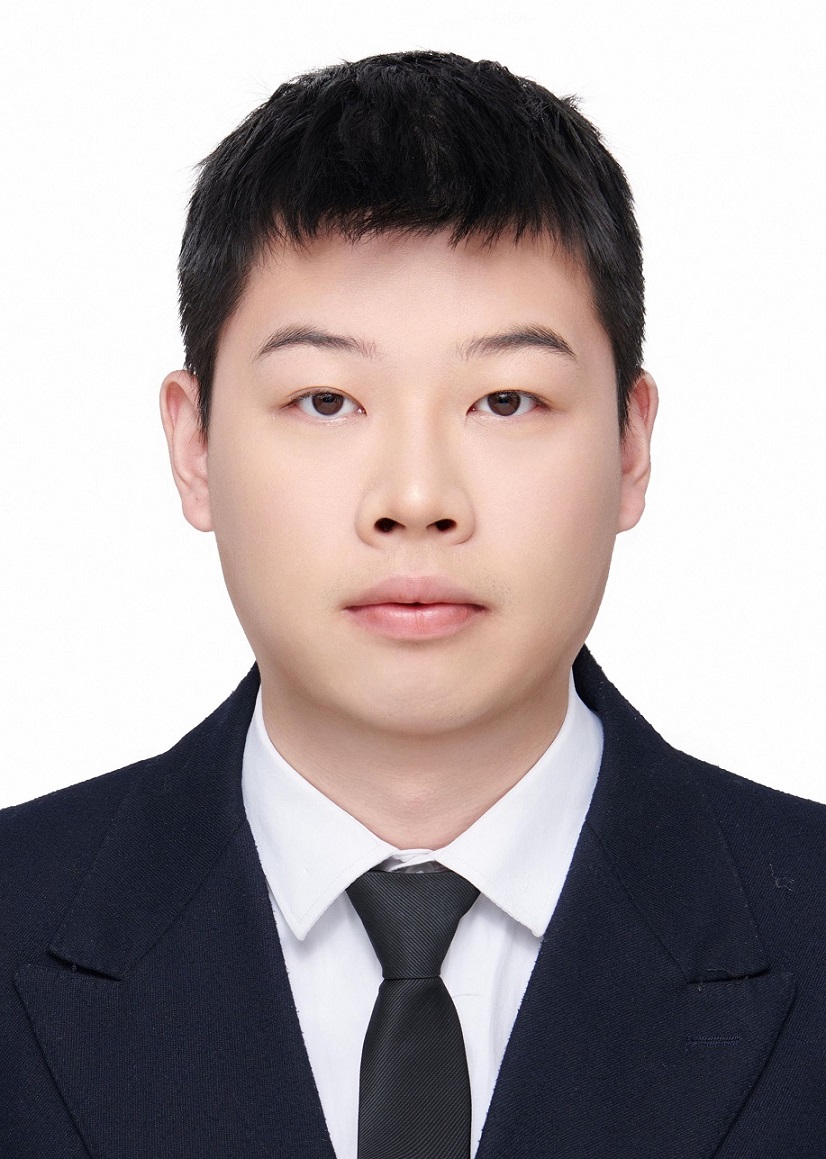}}]{Xin Liu} received the B.S. degree in computer science and technology from Huazhong University of Science and Technology, Wuhan, China, in 2022. He is pursuing the M.S. degree with the School of Computer Science and Technology, Huazhong University of Science and Technology, Wuhan, China. His interests include deep learning and adversarial  machine learning.
\end{IEEEbiography}

\begin{IEEEbiography}[{\includegraphics[width=1in,height=1.25in,clip,keepaspectratio]{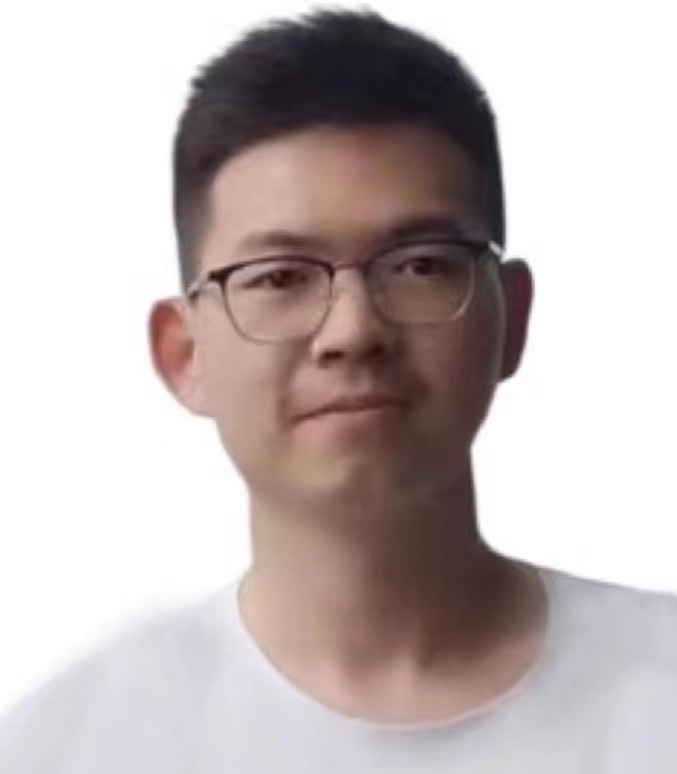}}]{Yichen Yang} received the B.S. degree in computer science and technology from Huazhong University of Science and Technology, Wuhan, China, in 2020. He is pursuing the M.S. degree with the School of Computer Science and Technology, Huazhong University of Science and Technology, Wuhan, China. His interests include deep learning and adversarial machine learning.
\end{IEEEbiography}

\begin{IEEEbiography}[{\includegraphics[width=1in,height=1.25in,clip,keepaspectratio]{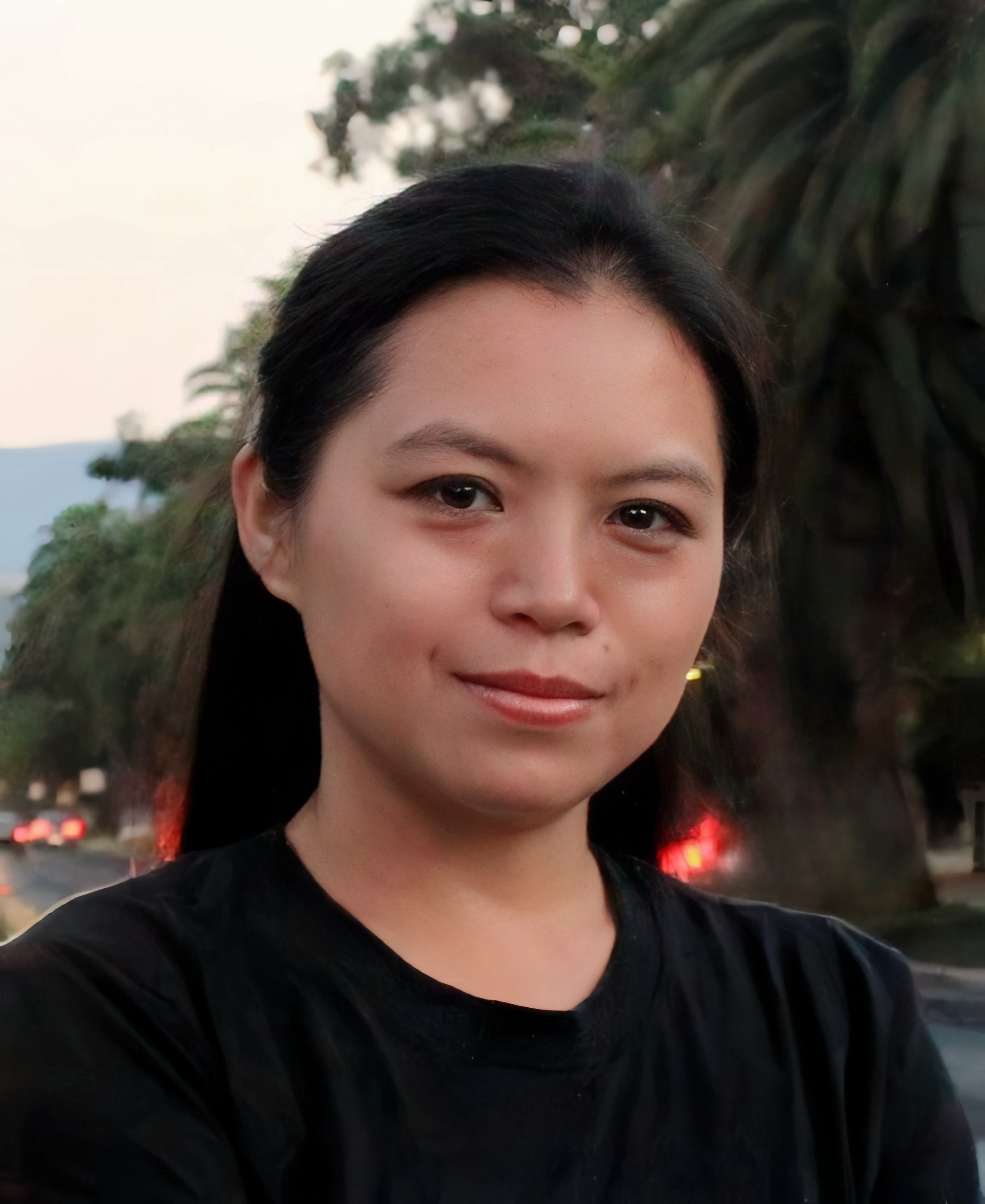}}]{Kun He}
(SM18) is currently a Professor in School of Computer Science and Technology, Huazhong University of Science and Technology, Wuhan, P.R. China. She received her Ph.D. in system engineering from Huazhong University of Science and Technology, Wuhan, China, in 2006. She had been with the Department of Management Science and Engineering at Stanford University in 2011-2012 as a visiting researcher. She had been with the department of Computer Science at Cornell University in 2013-2015 as a visiting associate professor, in 
2016 as a visiting professor, and in 2018 as a visiting professor. She was honored as a Mary Shepard B. Upson visiting professor for the 2016-2017 Academic year in Engineering, Cornell University, New York. Her research interests include adversarial machine learning, deep learning, graph data mining, and combinatorial optimization. 
\end{IEEEbiography}

\begin{IEEEbiography}[{\includegraphics[width=1in,height=1.25in,clip,keepaspectratio]{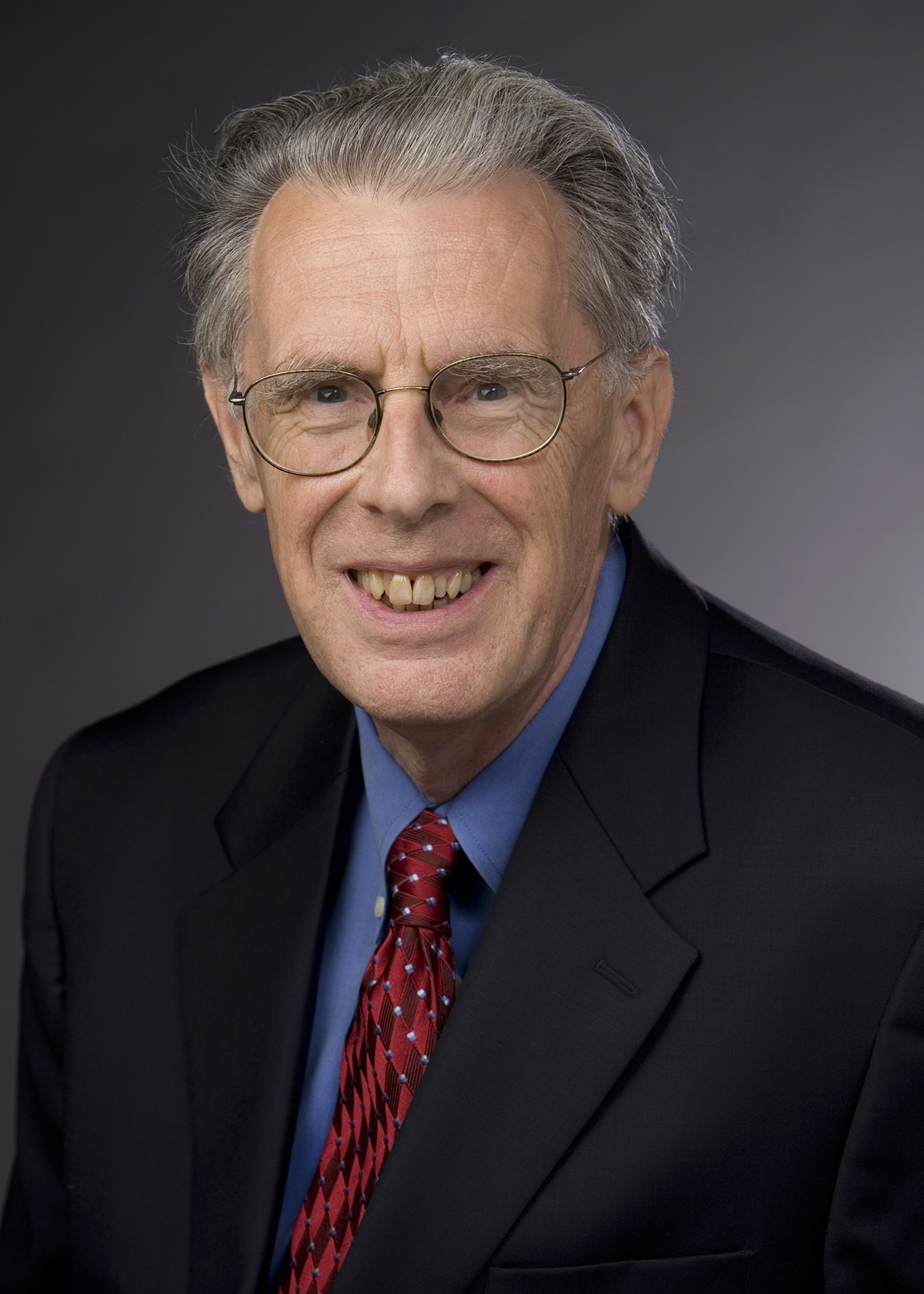}}]{John E. Hopcroft} (Fellow 1987, Life Fellow 2004) is the IBM Professor of Engineering and Applied Mathematics in Computer Science at Cornell University. 
  After receiving both his M.S. (1962) and Ph.D. (1964) in electrical engineering from Stanford University, he spent three years on the faculty of Princeton University. He joined the Cornell faculty in 1967, was named professor in 1972 and the Joseph C. Ford Professor of Computer Science in 1985.  He was honored with the A. M. Turing Award in 1986. 
  He is a member of the National Academy of Sciences (NAS), the National Academy of Engineering (NAE), a foreign member of the Chinese Academy of Sciences, and a fellow of the American Academy of Arts and Sciences (AAAS), the American Association for the Advancement of Science, the Institute of Electrical and Electronics Engineers (IEEE), and the Association of Computing Machinery (ACM).
  Hopcroft's research centers on theoretical aspects of computing, especially analysis of algorithms, automata theory, and graph algorithms. His most recent work is on the study of information capture and access.
  \end{IEEEbiography}

\end{document}